\documentclass{article} 
\usepackage{iclr2027_conference,times}

\usepackage{amsmath,amsfonts,bm}

\def\eqref#1{equation~\ref{#1}}

\def\1{\bm{1}}

\DeclareMathAlphabet{\mathsfit}{\encodingdefault}{\sfdefault}{m}{sl}
\SetMathAlphabet{\mathsfit}{bold}{\encodingdefault}{\sfdefault}{bx}{n}

\usepackage{hyperref}
\usepackage{url}

\usepackage{booktabs}
\usepackage{threeparttable}
\usepackage{wrapfig}
\usepackage{tabularx}
\usepackage{array}
\usepackage{graphicx}

\usepackage{multirow}
\usepackage{multicol}
\usepackage{makecell}
\usepackage{amssymb}  
\usepackage{bbding} 

\usepackage{algorithm}
\usepackage{algorithmic}

\usepackage{subcaption}

\def\ours{PlanContext}

\title{PlanContext: Benchmarking In-Context Driving Decision-Making}

\author{
    \textbf{Ruikai Li}\textsuperscript{\rm 1,2},
    \textbf{Naiyu Gao}\textsuperscript{\rm 2},
    \textbf{Hao Shan}\textsuperscript{\rm 1},
    \textbf{Yizhe Fan}\textsuperscript{\rm 1},
    \textbf{Yanhao Wu}\textsuperscript{\rm 2}, 
    \textbf{Han Jiang}\textsuperscript{\rm 1}, 
    \textbf{Yilong Ren}\textsuperscript{\rm 1}, \\ 
    \textbf{Zhenning Li}\textsuperscript{\rm 3},
    \textbf{Haiyang Yu}\textsuperscript{\rm 1}, 
    \textbf{Yanhu Shan}\textsuperscript{\rm 2}, 
    \textbf{Rui Wu}\textsuperscript{\rm 2}, 
    \textbf{Hao Zhao}\textsuperscript{\rm 4,5},  
    \textbf{Hongyang Li}\textsuperscript{\rm 6}, 
    \textbf{Zhiyong Cui}\textsuperscript{\rm 1}$^{\dagger}$\\
    \textsuperscript{\rm 1} Beihang University \qquad
    \textsuperscript{\rm 2} Horizon Robotics \qquad
    \textsuperscript{\rm 3} University of Macau \qquad
    \textsuperscript{\rm 4} BAAI \qquad \\
    \textsuperscript{\rm 5} AIR, Tsinghua University \qquad
    \textsuperscript{\rm 6} OpenDriveLab, The University of Hong Kong \\ 
    {\small $^{\dagger}$Corresponding author: zhiyongc@buaa.edu.cn
    \qquad Project:\href{https://plancontext.github.io/}{https://plancontext.github.io/}}
}

\iclrfinalcopy 
\begin{document}

\maketitle

\begin{abstract}
Real-world driving requires context-aware decision-making. Existing end-to-end driving benchmarks primarily assess trajectory quality through safety, compliance, comfort, and driving progress, leaving context-dependent decision-making insufficiently characterized. To fill this gap, we present \textbf{PlanContext}, a benchmark for evaluating \textbf{in-context driving decision-making}. PlanContext constructs paired evaluations by varying high-level driving intentions while keeping the observed scene fixed, enabling a natural and controllable assessment of context-dependent decision-making. We further define six representative driving contexts and jointly evaluate contextual decision correctness, road compliance, and dynamic safety with class-balanced metrics. Experiments on 9,561 scene--condition instances from NAVSIM and 23 public end-to-end driving models show that high conventional driving-quality scores do not consistently correspond to strong route-level in-context driving decision-making. Further analysis shows that model behavior varies with decision timing, road structure, and traffic interaction depending on the decision required by the current context. Our results highlight in-context driving decision-making as an important complementary dimension for evaluating end-to-end driving systems. 
\end{abstract}

\vspace{-5mm}
\begin{figure}[htbp]
    \centering
    \includegraphics[width=1.0\linewidth]{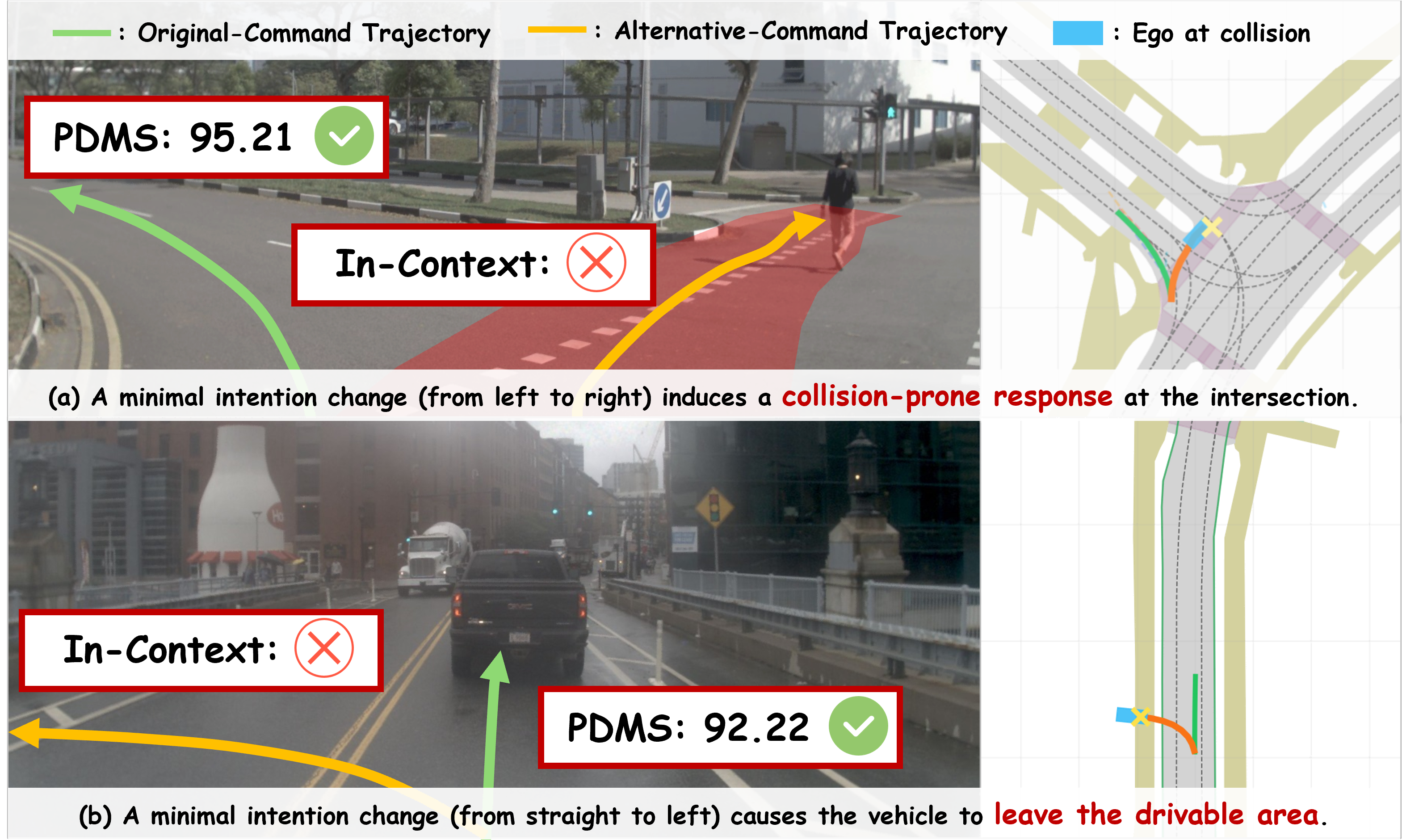}
    \vspace{-5mm}
    \caption{\textbf{\ours.} Conventional driving metrics primarily evaluate trajectory quality in terms of safety, compliance, comfort, and driving progress. However, models with high PDMS can still produce context-inappropriate trajectories when navigation intentions change. \ours\  uses a natural and controllable navigation intervention to evaluate in-context driving decision-making.}
    \label{fig:teaser}
\end{figure}


\section{Introduction}
\label{sec:1-Introduction}

\vspace{-3mm}

Autonomous driving has attracted growing attention from both academia and industry due to its potential to improve road safety and transportation efficiency~\citep{chen2024end,janai2020computer}. The community has developed a wide range of end-to-end driving models that integrate sensory observations, ego states, and high-level driving intentions to directly predict future trajectories or driving actions~\citep{chitta2022transfuser,hu2023uniad,zhou2026qwendrive}. As model families and technical paradigms continue to diversify, reliable evaluation and fair comparison of driving performance have become increasingly important for measuring progress in the field.

However, comprehensively characterizing the decision-making capabilities of end-to-end driving models remains challenging. Existing benchmarks primarily assess trajectory quality through metrics such as collision avoidance, road compliance, comfort, and driving progress~\citep{caesar2020nuscenes,jia2024bench2drive,cao2025navsimv2,zhang2026plug}. While these metrics are effective for measuring whether a model produces safe and reasonable trajectories, a high-quality trajectory alone does not establish that the model has correctly interpreted the current driving context and made an appropriate decision.  As shown in Fig.~\ref{fig:teaser}, models with high PDMS under the original navigation command can still respond inappropriately when the command changes. In Fig.~\ref{fig:teaser}(a), the vehicle has already started a left-turn maneuver, and a newly issued right-turn command should not trigger an immediate reversal that drives the vehicle toward a pedestrian. Similarly, in Fig.~\ref{fig:teaser}(b), when a turning instruction is given before the corresponding maneuver becomes locally feasible, the policy should continue to follow the current road rather than turn immediately and leave the drivable area. These cases highlight that driving decisions should be made in accordance with the current driving context. A reliable autonomous driving policy should jointly consider contextual information such as the current vehicle state, road geometry, and high-level navigation intention to make an appropriate decision. Yet existing benchmarks offer limited insight into such in-context decision-making capability, allowing context-inappropriate decision failures to remain hidden behind strong trajectory-level performance.


To address this gap, we introduce \textbf{\ours}, a benchmark for systematically evaluating \textbf{in-context driving decision-making} in end-to-end autonomous driving. \ours\ keeps sensor observations and vehicle history fixed while applying controlled changes to high-level navigation intentions, enabling paired comparison of model decisions under the same environmental context. Based on road topology, target reachability, decision timing, and ongoing vehicle motion, we construct representative driving contexts and define context-specific requirements for whether an intention should be executed, deferred, rejected, or preserved. \ours\ jointly evaluates contextual decision correctness, road compliance, and dynamic safety, and further analyzes model behavior under different contextual factors. In this way, \ours\ extends end-to-end driving evaluation beyond trajectory quality toward a systematic assessment of intention--context--decision consistency.

Based on \ours, we construct 9,561 strictly filtered scene--condition instances from NAVSIM~\citep{dauner2024navsim} and evaluate 23 publicly available end-to-end driving models spanning three paradigms: vision/state-to-action, vision-language-action, and world-model-based action. Model rankings under conventional driving-quality evaluation differ substantially from those under route-level in-context evaluation. Across all 23 models, contextual success reaches only 22.46\% at best when the driving context requires executing an immediately actionable alternative route. Moreover, no consistent performance trend emerges across model paradigms: VA, VLA, and WA models all exhibit substantial within-paradigm variation in in-context decision-making performance. Our key contributions are threefold:

\vspace{-2mm}
\begin{itemize}
\item \textbf{We introduce \ours, a benchmark for in-context driving decision-making with 9,561 curated scene--condition instances.} It enables systematic evaluation of whether end-to-end models make appropriate decisions by jointly considering driving context and high-level intentions.

\item \textbf{\ours\ provides evaluation criteria and metrics for in-context driving decision-making across six scenario categories.} The evaluation jointly assesses contextual decision correctness, road compliance, and dynamic safety, with separate metrics for route-level and lane-level performance.

\item \textbf{Our evaluation of 23 end-to-end driving models shows that conventional driving-quality metrics do not fully capture in-context decision-making capability.} Further analyses reveal category-dependent performance patterns across decision timing, road structure, and traffic interactions.

\end{itemize}

\vspace{-2mm}
\section{Related Work}
\label{sec:2-RelatedWork}

\vspace{-3mm}
\paragraph{End-to-End Autonomous Driving.} End-to-end autonomous driving directly maps sensory observations and vehicle states to future trajectories or actions. Recent approaches span perception–planning architectures, direct trajectory regression, generative planners, vision–language–action models, and world-model-based driving systems~\citep{jiang2023vad,chitta2022transfuser,liao2025diffusiondrive,zhou2026autovla,wang2026drive-jepa}. Beyond sensory inputs, modern models increasingly incorporate navigation goals and structured scene information into decision-making. Recent work further shows that decision-relevant traffic elements, including traffic lights, road signs, and lane topology, consistently affect planning performance across diverse end-to-end paradigms~\citep{zhang2026plug}. These developments motivate evaluation beyond aggregate trajectory quality toward how models use contextual information when making driving decisions.

\vspace{-3mm}
\paragraph{Evaluation of End-to-End Driving.} End-to-end driving evaluation has evolved from open-loop trajectory similarity toward more comprehensive measurements of driving quality and robustness. NAVSIM~\citep{dauner2024navsim} introduces data-driven non-reactive simulation and evaluates predicted trajectories through collision, road compliance, comfort, time-to-collision, and driving progress. Closed-loop benchmarks such as Bench2Drive~\citep{jia2024bench2drive} further evaluate interactive driving performance, while NAVSIMv2~\citep{cao2025navsimv2} augments logged data with synthetic observations to assess recovery from ego deviations. Fail2Drive~\citep{gerstenecker2026fail2drive} further examines closed-loop generalization by pairing distribution-shifted routes with in-distribution counterparts under changes in appearance, layout, and agent behavior. These benchmarks substantially improve the evaluation of trajectory quality, safety, and robustness. \ours\ provides a complementary evaluation perspective by examining whether driving decisions remain appropriate under different combinations of high-level intentions and environmental contexts, thereby revealing aspects of decision-making that are not captured by conventional driving-quality metrics.

\vspace{-3mm}
\paragraph{Instruction-Conditioned and Intervention-Based Evaluation.} High-level navigation commands help resolve behavioral ambiguity in end-to-end driving~\citep{codevilla2018conditionalimitation,chitta2022transfuser}. NavControl~\citep{wu2025navcontrol} constructs feasible alternative commands in NAVSIM and evaluates navigation compliance alongside driving quality. ICR-Drive~\citep{hamid2026icrdrive} assesses robustness to goal-preserving and goal-conflicting language perturbations on matched CARLA routes, using the original navigation objective as the evaluation reference. Beyond driving policies, What-If World~\citep{cai2026whatifworld} evaluates whether video world models produce the expected physical outcomes under paired prompts differing in a single physical variable. \ours{} targets context-dependent decision correctness through paired intention changes with observations and vehicle history fixed. Its criteria account for road topology, target reachability, decision timing, and ongoing motion, assessing both the semantic validity of individual predictions and whether paired predictions exhibit context-appropriate behavioral adjustment or preservation.

\vspace{-2mm}
\section{\ours: Evaluating In-Context Driving Decision-Making}
\label{sec:3-PlanContext}

\vspace{-2mm}

\ours\ evaluates in-context driving decision-making through paired navigation interventions and context-aware assessment. We first introduce the intervention protocol and the construction of representative driving contexts in Sections~\ref{subsec:3-1-paired_intervention} and~\ref{subsec:3-2-scenario_construction}. Section~\ref{subsec:3-3-contextual_evaluation} then defines the contextual decision evaluation criteria, followed by the class-balanced performance metrics in Section~\ref{subsec:3-4-balanced_metrics}.

\vspace{-2mm}
\subsection{Paired Navigation Intervention}
\label{subsec:3-1-paired_intervention}
\vspace{-2mm}

Given a driving scene $s_i$, its original navigation intention $c_i^{N}$, and an alternative intention $c_i^{F}$, an evaluated model $f_{\theta}$ predicts two future trajectories:
\begin{equation}
T_i^{N}=f_{\theta}(s_i,c_i^{N}), \qquad
T_i^{F}=f_{\theta}(s_i,c_i^{F}).
\end{equation}
Each trajectory contains eight future ego poses over a 4-second horizon. \ours\ keeps the sensor observations and vehicle history in $s_i$ unchanged and only modifies the high-level navigation intention, yielding paired predictions under the same environmental context.

For each condition $x\in\{N,F\}$, let $\mathcal{A}(s_i,c_i^{x})$ denote the set of behaviors that are semantically valid under scene $s_i$ and intention $c_i^{x}$. The semantic-validity indicator is defined as
\begin{equation}
S_i^{x}
=
\mathbb{1}
\left[
T_i^{x}\in\mathcal{A}(s_i,c_i^{x})
\right],
\qquad
x\in\{N,F\}.
\end{equation}
This formulation allows multiple valid future behaviors under the same intention and avoids treating a single recorded trajectory as the unique correct outcome.

The logged trajectory $T_i^{GT}$ is used to characterize the natural road evolution, maneuver stage, and route-commitment state of the scene. These cues support the construction of the context-dependent valid behavior sets $\mathcal{A}(s_i,c_i^{N})$ and $\mathcal{A}(s_i,c_i^{F})$. The paired relationship between $T_i^{N}$ and $T_i^{F}$ is evaluated separately in Section~\ref{subsec:3-3-contextual_evaluation}.

\vspace{-2mm}
\subsection{Context-Aware Scenario Construction}
\label{subsec:3-2-scenario_construction}
\vspace{-2mm}

\begin{figure*}[t]
    \centering
    \includegraphics[width=0.98\textwidth]{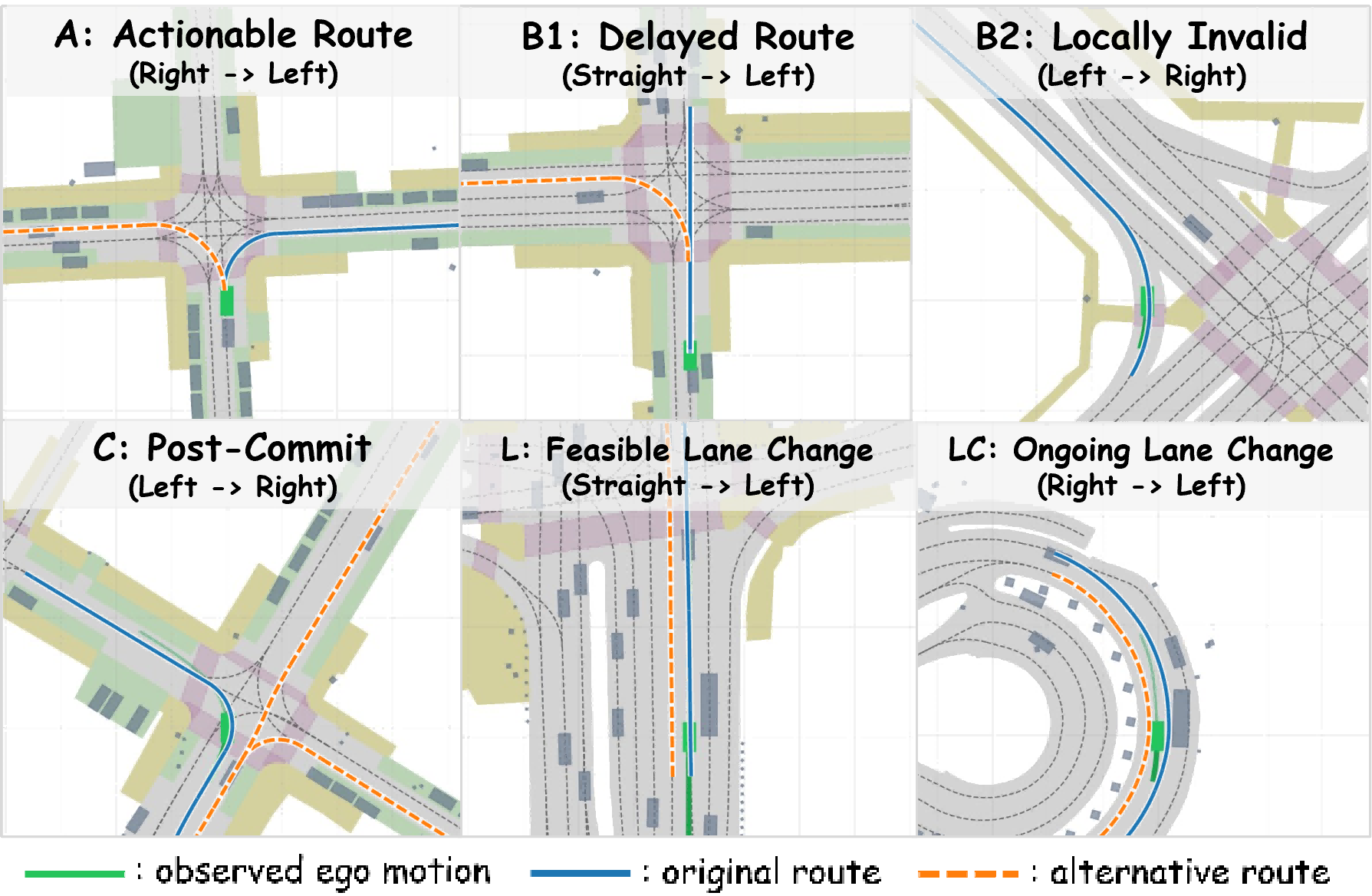}
    \vspace{-2mm}
    \caption{
    Representative examples of the six \ours\ categories. Parentheses indicate the change in navigation intention.
    }
    \label{fig:plancontext_categories}
    \vspace{-6mm}
\end{figure*}

The valid behavior sets introduced in Section~\ref{subsec:3-1-paired_intervention} are determined by the driving context of each scene--condition instance. We characterize this context using road topology, target reachability, decision timing, and the ongoing motion state of the ego vehicle. Together, these factors determine how a high-level driving intention can be appropriately realized in the current scene.

\begin{wraptable}{r}{0.49\textwidth}
\vspace{-5mm}
\centering
\caption{\ours\ scenario categories and contextual requirements.}
\label{tab:plancontext_categories}
\vspace{-3mm}
\scriptsize
\setlength{\tabcolsep}{3pt}
\renewcommand{\arraystretch}{1.08}

\begin{tabularx}{\linewidth}{
    c
    >{\raggedright\arraybackslash}X
    l
}
\toprule
Category & Context condition & Requirement \\
\midrule
A  & Target route reachable within 2\,s       & Follow \\
B1 & Target route reachable in 4.5--8\,s      & Wait \\
B2 & No locally feasible target               & Reject \\
C  & Ego beyond route commitment point        & Reject \\
L  & Adjacent target lane geometrically and topologically reachable   & Follow \\
LC & Lateral maneuver already in progress     & Preserve \\
\bottomrule
\end{tabularx}
\vspace{-3mm}
\end{wraptable}

\ours\ organizes the resulting instances into four contextual requirements:
\emph{follow}, \emph{wait}, \emph{reject}, and \emph{preserve}.
\emph{Follow} applies when the alternative target is currently reachable and should be executed within the evaluation horizon.
\emph{Wait} describes cases where the target becomes feasible in the near future while the current state still favors maintaining the existing route.
\emph{Reject} covers locally unreachable targets or cases where the vehicle has already passed the appropriate decision point.
\emph{Preserve} applies when an ongoing maneuver should remain temporally consistent after the intention changes.
We instantiate these requirements with six complementary scenario categories, summarized in Table~\ref{tab:plancontext_categories} and illustrated in Fig.~\ref{fig:plancontext_categories}.

The route-level categories A, B1, B2, and C capture different stages of navigation feasibility and route commitment. A represents an immediately actionable route change, while B1 contains targets that become feasible beyond the current decision window. B2 captures locally unsupported intentions, and C describes scenes in which the ego vehicle has already committed to the original route. The lane-level categories complement these route-level contexts: L evaluates feasible adjacent-lane transitions, while LC evaluates the continuity of an ongoing lateral maneuver.

For A, B1, B2, C, and L, each category defines the semantic behavior sets $\mathcal{A}(s_i,c_i^{N})$ and $\mathcal{A}(s_i,c_i^{F})$
used to compute $S_i^{N}$ and $S_i^{F}$ in Section~\ref{subsec:3-1-paired_intervention}. LC instead defines a continuity requirement for the ongoing lateral maneuver, as detailed in Section~\ref{subsec:3-3-contextual_evaluation}.
We retain only scene--condition instances with sufficiently clear contextual evidence for reliable category assignment. Detailed scenario mining, filtering, and category assignment criteria are provided in Sec.~\ref{sec:appendix_scenario}.

To further assess benchmark validity, three reviewers with backgrounds in autonomous driving independently inspected a randomly sampled subset of 500 scene--condition instances and 11,500 model-instance evaluations from all 23 evaluated models after the evaluation rules were finalized. The reviewers were blinded to the automated evaluation results during independent assessment. Following discussion, approximately 0.7\% of the inspected model-instance evaluations were judged inconsistent with the automated evaluation outcomes. Detailed findings are provided in Appendix~\ref{sec:appendix_human_validation}.

\vspace{-2mm}
\subsection{Contextual Decision Evaluation}
\label{subsec:3-3-contextual_evaluation}
\vspace{-2mm}

Given the contextual requirement assigned in Section~\ref{subsec:3-2-scenario_construction}, \ours\ evaluates each scene--condition instance through three components: contextual decision correctness, road compliance, and dynamic safety.

\paragraph{Paired behavioral correctness.}
We first characterize how the prediction changes when the navigation intention changes. The original and alternative trajectories are projected into a road-relative coordinate system, where longitudinal progress, lateral displacement, and heading deviation are measured with respect to the local road geometry. The trajectories are aligned by road progress to reduce the influence of speed differences on behavioral comparison.

Let $k_i\in\{A,B1,B2,C,L,LC\}$ denote the \ours\ category of instance $i$. We define a category-specific paired behavior function
\begin{equation}
P_i
=
\Phi_{k_i}
\left(
T_i^{N},T_i^{F},s_i
\right),
\end{equation}
where $\Phi_{k_i}$ evaluates whether the behavioral change between the original and alternative predictions satisfies the requirement of category $k_i$.

For the \emph{follow} categories A and L, $P_i$ requires an identifiable path change toward the alternative target. For B1, B2, and C, it requires the alternative intention to avoid inducing an inappropriate target-directed deviation. LC is treated separately because its objective is to preserve the continuity of an ongoing lateral maneuver rather than to evaluate two independent semantic outcomes. We therefore assign LC predictions a continuity status \(P_i^{LC}\in\{\mathrm{correct},\mathrm{incorrect},\mathrm{uncertain}\}\). Detailed category-specific behavioral criteria and trajectory comparison rules are provided in Sec.~\ref{sec:appendix_evaluation}.

For A, B1, B2, C, and L, contextual decision correctness combines the semantic validity of both predictions with their paired behavioral relationship:
\begin{equation}
C_i
=
\begin{cases}
S_i^{N}
\land
S_i^{F}
\land
P_i, & k_i \in \{A,B1,B2,C,L\}, \\[4pt]
1, & k_i = LC,\ P_i^{LC}=\mathrm{correct}, \\[4pt]
0, & k_i = LC,\ P_i^{LC}=\mathrm{incorrect}.
\end{cases}
\end{equation}
LC instances with \(P_i^{LC}=\mathrm{uncertain}\) are excluded from subsequent success-rate aggregation.

\paragraph{Road compliance.}
Road compliance is evaluated using the full ego-vehicle footprint. Let $\mathcal{B}(T_{i,t}^{F})$ denote the vehicle footprint at future step $t$ and $\Omega_i^{F}$ the road region permitted by the corresponding context. The road-compliance indicator is
\begin{equation}
R_i^{F}
=
\prod_{t=1}^{8}
\mathbb{1}
\left[
\mathcal{B}(T_{i,t}^{F}) \subseteq \Omega_i^{F}
\right].
\end{equation}

\paragraph{Dynamic safety.}
Dynamic safety is evaluated using the NAVSIM safety criteria, including collision, drivable-area compliance, wrong-way driving, traffic-light violations, time-to-collision, and comfort. We denote the resulting safety indicator as $E_i^{F}$.

The final contextual success for instance $i$ is defined as
\begin{equation}
q_i
=
C_i
\land
R_i^{F}
\land
E_i^{F}.
\end{equation}

Thus, a successful instance requires a context-consistent decision response, road-compliant motion, and dynamic safety. For A, B1, B2, C, and L, contextual correctness additionally requires semantic validity under both navigation conditions.

\subsection{Class-Balanced \ours\ Metrics}
\label{subsec:3-4-balanced_metrics}

The six \ours\ categories are highly imbalanced in sample size, making a direct micro-average sensitive to the empirical category distribution. We therefore first compute the contextual success rate independently for each category:
\begin{equation}
S_k
=
\frac{1}{|\mathcal{D}_k|}
\sum_{i\in\mathcal{D}_k} q_i,
\qquad
k \in \{A,B1,B2,C,L,LC\},
\end{equation}
where $\mathcal{D}_k$ denotes the set of valid scene--condition instances used for metric aggregation in category $k$. LC instances labeled as uncertain are excluded.

We aggregate the scores of A, B1, B2, and C to evaluate route-level in-context driving decision-making. The scores of L and LC are reported separately to evaluate lane-level in-context driving decision-making.

The Navigation Action Score (NAS) is the contextual success rate in category A, where the alternative route is immediately actionable.
The Context Preservation Score (CPS) measures contextual success when the alternative navigation intention should be deferred or rejected, averaged across categories B1, B2, and C:
\begin{equation}
\mathrm{CPS}
=
\frac{S_{B1}+S_{B2}+S_C}{3}.
\end{equation}

We define the Navigation Context Score (NCS) as the geometric mean of NAS and CPS to summarize route-level in-context driving decision-making:
\begin{equation}
\mathrm{NCS}
=
\sqrt{\mathrm{NAS}\cdot\mathrm{CPS}}.
\end{equation}

The geometric mean penalizes imbalanced performance across NAS and CPS.

For lane-level evaluation, we report the Lane Adaptation Score (LAS) and Maneuver Continuity Score (MCS), defined as the contextual success rates in categories L and LC, respectively. We report all five metrics together with the six category-wise scores.

\vspace{-2mm}
\section{Experiments}
\label{sec:4-exp}
\vspace{-2mm}

\subsection{Experimental Setup}
\label{subsec:4-1-experimental_setup}

\paragraph{Benchmark.}
\ours\ is constructed from the NAVSIM \texttt{navtest} split. After scenario mining and context-aware filtering, the benchmark contains 9,561 scene--condition instances from 5,911 unique physical scenes. The six contextual categories include 334 A, 251 B1, 5,285 B2, 580 C, 2,716 L, and 395 LC instances. For each retained instance, the sensor observations and vehicle history are fixed, while the evaluated model is queried under both the original navigation intention and the corresponding alternative intention.

\begin{table*}[t]
\centering
\caption{Overall comparison of 23 end-to-end driving models on \ours.}
\label{tab:main_results}

\small
\setlength{\tabcolsep}{1.5pt}
\setlength{\abovecaptionskip}{0.1cm}

\begin{tabular}{llrrrrrrr}
\toprule
Model & Venue & PDMS & mPDMS & NAS & CPS & NCS & LAS & MCS \\
\midrule

\multicolumn{9}{l}{\textbf{VA Models}} \\
DrivoR~\citep{kirby2026drivor} & CVPR 2026
& 93.96 & 83.56 & 19.76 & 61.23 & 34.78 & 0.00 & 92.62 \\
GoalFlow~\citep{xing2025goalflow} & CVPR 2025
& 88.31 & 79.82 & 15.57 & 63.89 & 31.54 & 0.07 & 87.82 \\
TransFuser~\citep{chitta2022transfuser} & IEEE TPAMI
& 83.39 & 74.16 & 13.77 & 52.98 & 27.01 & 0.15 & 73.59 \\
SparseDriveV2~\citep{sun2026sparsedrivev2} & ECCV 2026
& 92.22 & 81.22 & 12.57 & 56.25 & 26.60 & 0.15 & 93.64 \\
DiffusionDrive~\citep{liao2025diffusiondrive} & CVPR 2025
& 88.05 & 79.80 & 3.29 & 65.44 & 14.68 & 0.00 & 92.11 \\
StateMLP~\citep{dauner2024navsim} & NeurIPS 2024
& 65.52 & 56.95 & 15.57 & 8.68 & 11.63 & \textbf{8.73} & 31.55 \\
CLOVER~\citep{ang2026clover} & Arxiv 2026
& 94.48 & 94.35 & 0.60 & 57.35 & 5.86 & 0.22 & 85.82 \\
DiffusionDriveV2~\citep{zou2025diffusiondrivev2} & Arxiv 2026
& 90.95 & 90.89 & 0.00 & 49.51 & 0.00 & 0.07 & 82.19 \\

\midrule
\multicolumn{9}{l}{\textbf{VLA Models}} \\
OneVL~\citep{lu2026onevl} & Arxiv 2026
& 88.92 & 80.79 & \textbf{22.46} & 68.79 & \textbf{39.30} & 0.00 & 91.88 \\
ChainFlow-VLA~\citep{wang2026chainflow} & Arxiv 2026
& 95.21 & 83.14 & 20.96 & 58.88 & 35.13 & 0.04 & 94.42 \\
ReCogDrive-IL~\citep{xiong2026recogdrive} & ICLR 2026
& 86.00 & 78.59 & 21.26 & 55.39 & 34.31 & 0.00 & 82.56 \\
ReCogDrive-RL~\citep{xiong2026recogdrive} & ICLR 2026
& 90.46 & 78.93 & 18.56 & 53.93 & 31.64 & 0.07 & 84.83 \\
NoRD~\citep{rawal2026nord} & CVPR 2026
& 85.69 & 75.75 & 17.66 & 49.31 & 29.51 & 0.18 & 82.69 \\
AutoVLA~\citep{zhou2026autovla} & NeurIPS 2025
& 92.06 & 80.92 & 12.28 & 59.76 & 27.08 & 0.11 & 86.67 \\
DriveVLA-W0~\citep{li2026drivevla} & ICLR 2026
& 87.02 & 78.54 & 4.49 & 63.01 & 16.82 & 0.00 & 91.88 \\
DP-VLA~\citep{zheng2026dp-vla} & Arxiv 2026
& 88.56 & 88.41 & 1.80 & 66.09 & 10.90 & 0.00 & 86.58 \\
Curious-VLA~\citep{chen2026Curious-VLA} & CVPR 2026
& 89.20 & 89.32 & 0.60 & 60.62 & 6.02 & 0.15 & 87.95 \\
WAM-Flow~\citep{xu2026wam-flow} & CVPR 2026
& 90.27 & 90.20 & 0.60 & 57.27 & 5.86 & 0.00 & 88.27 \\

\midrule
\multicolumn{9}{l}{\textbf{WA Models}} \\
SUV~\citep{yuan2026suv} & Arxiv 2026
& 90.87 & 81.63 & 14.67 & \textbf{69.94} & 32.03 & 0.00 & \textbf{94.68} \\
WoTE~\citep{li2025wote} & ICCV 2025
& 88.22 & 79.21 & 13.47 & 60.36 & 28.52 & 0.04 & 91.88 \\
SimWAM~\citep{zhao2026simwam} & Arxiv 2026
& 90.97 & 90.76 & 0.30 & 56.74 & 4.12 & 0.00 & 88.83 \\
Drive-JEPA~\citep{wang2026drive-jepa} & Arxiv 2026
& 93.69 & 94.18 & 0.00 & 52.37 & 0.00 & 0.33 & 85.71 \\
DynVLA Stage 2~\citep{shang2026dynvla} & ICML 2026
& 87.20 & 86.53 & 0.00 & 64.54 & 0.00 & 0.00 & 88.61 \\

\bottomrule
\end{tabular}

\vspace{2pt}

\begin{minipage}{0.98\textwidth}
\scriptsize
\textbullet\; PDMS denotes conventional driving quality on the full NAVSIM v1 \texttt{navtest} split.\\[-1pt]
\textbullet\; mPDMS denotes PDMS recomputed under the original navigation condition on the 5,911 physical scenes selected by \ours.\\[-1pt]
\textbullet\; NAS is the contextual success rate in category A.\\[-1pt]
\textbullet\; CPS is the macro-average of the success rates in categories B1, B2, and C.\\[-1pt]
\textbullet\; NCS is the geometric mean of NAS and CPS and summarizes route-level in-context driving decision-making.\\[-1pt]
\textbullet\; LAS and MCS are the contextual success rates in L and LC, respectively, for lane-level evaluation.\\[-1pt]
\textbullet\; The 95\% confidence intervals for NCS are reported in Table~\ref{tab:ncs_confidence_intervals}.
\end{minipage}

\end{table*}

\paragraph{Models.}
We evaluate 23 publicly available end-to-end driving models covering three representative paradigms. Vision/state-to-action (VA) models directly predict driving trajectories from sensory or state inputs, vision-language-action (VLA) models incorporate vision-language representations into planning, and world-model-based action (WA) models explicitly predict or encode future scene dynamics. All models are evaluated using publicly released or official assets, with navigation conditions explicitly involved in inference. Their outputs are standardized as eight future ego poses over a 4-second horizon.

\paragraph{Evaluation Protocol.}
For each model, we report its conventional driving performance on the full NAVSIM \texttt{navtest} split using PDMS. To enable a scene-matched comparison with \ours, we additionally compute matched PDMS (mPDMS) under the original navigation condition on the same 5,911 physical scenes selected by \ours. Each physical scene contributes once to mPDMS, even when it is associated with multiple alternative navigation conditions. \ours\ then performs paired inference under the original and alternative navigation intentions on the 9,561 scene--condition instances. We report NAS, CPS, and NCS for route-level in-context driving decision-making, and LAS and MCS for lane-level evaluation. We additionally report contextual success rates for all six categories.

\paragraph{Statistical Analysis.}
We estimate uncertainty using 10,000 bootstrap resamples over the 5,911 unique scene tokens. All scene--condition instances and model predictions associated with a sampled scene are retained together to preserve the paired evaluation structure. We report percentile-based 95\% confidence intervals for NCS. Pairwise model comparisons are computed on matched bootstrap samples, with Benjamini--Hochberg correction applied across all 253 pairwise comparisons.

\begin{figure}[t]
  \centering
  \captionsetup[subfigure]{skip=3pt}
  \begin{subfigure}[b]{0.32\textwidth}
    \centering
    \includegraphics[width=\linewidth,height=4.05cm,keepaspectratio]{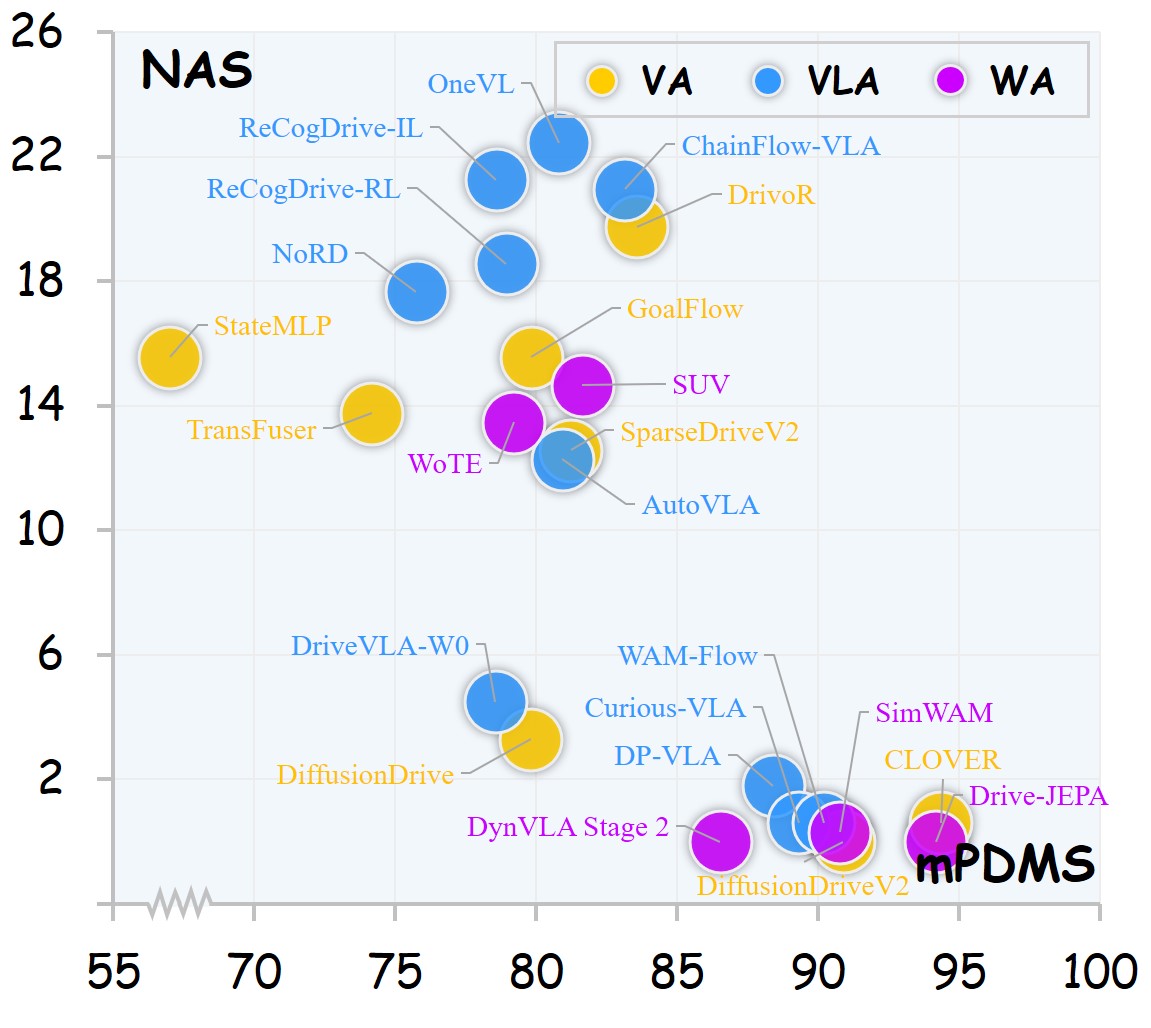}
    \caption{mPDMS \textit{vs}. NAS}
    \label{fig:rel_sub1}
  \end{subfigure}\hfill
  \begin{subfigure}[b]{0.32\textwidth}
    \centering
    \includegraphics[width=\linewidth,height=4.05cm,keepaspectratio]{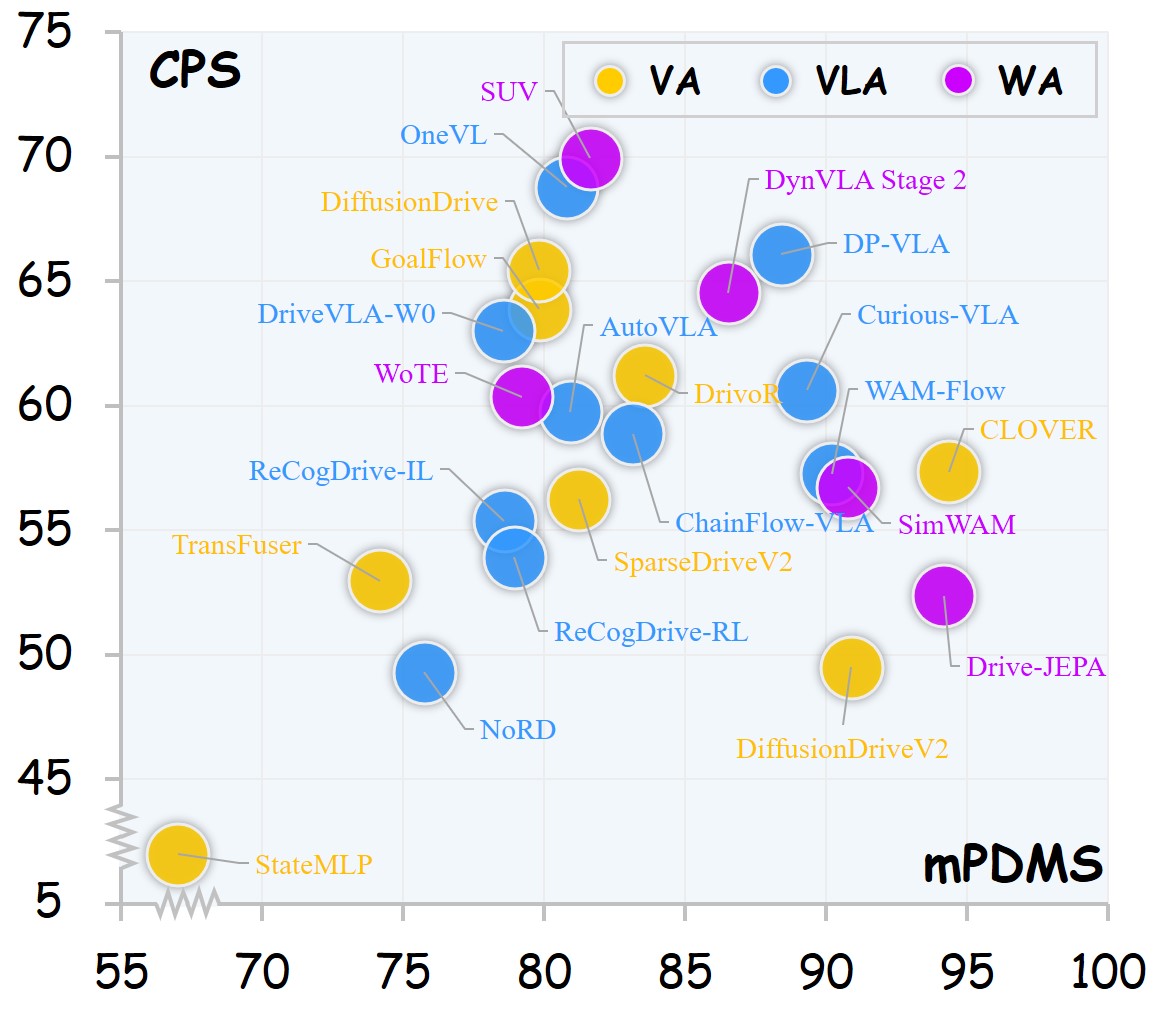}
    \caption{mPDMS \textit{vs}. CPS}
    \label{fig:rel_sub2}
  \end{subfigure}\hfill
  \begin{subfigure}[b]{0.32\textwidth}
    \centering
    \includegraphics[width=\linewidth,height=4.05cm,keepaspectratio]{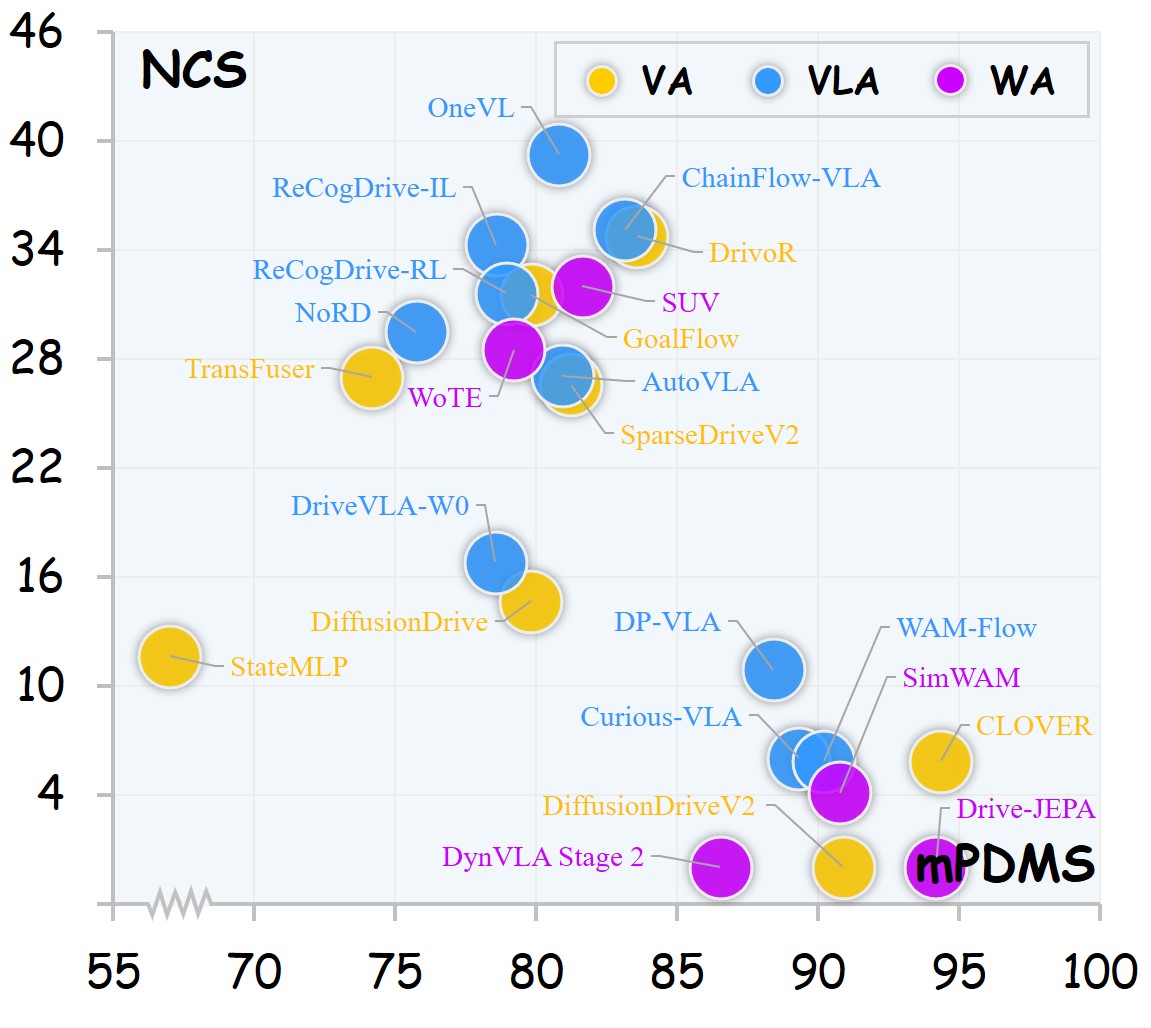}
    \caption{mPDMS \textit{vs}. NCS}
    \label{fig:rel_sub3}
  \end{subfigure}

  \medskip
  \begin{subfigure}[b]{0.32\textwidth}
    \centering
    \includegraphics[width=\linewidth,height=4.05cm,keepaspectratio]{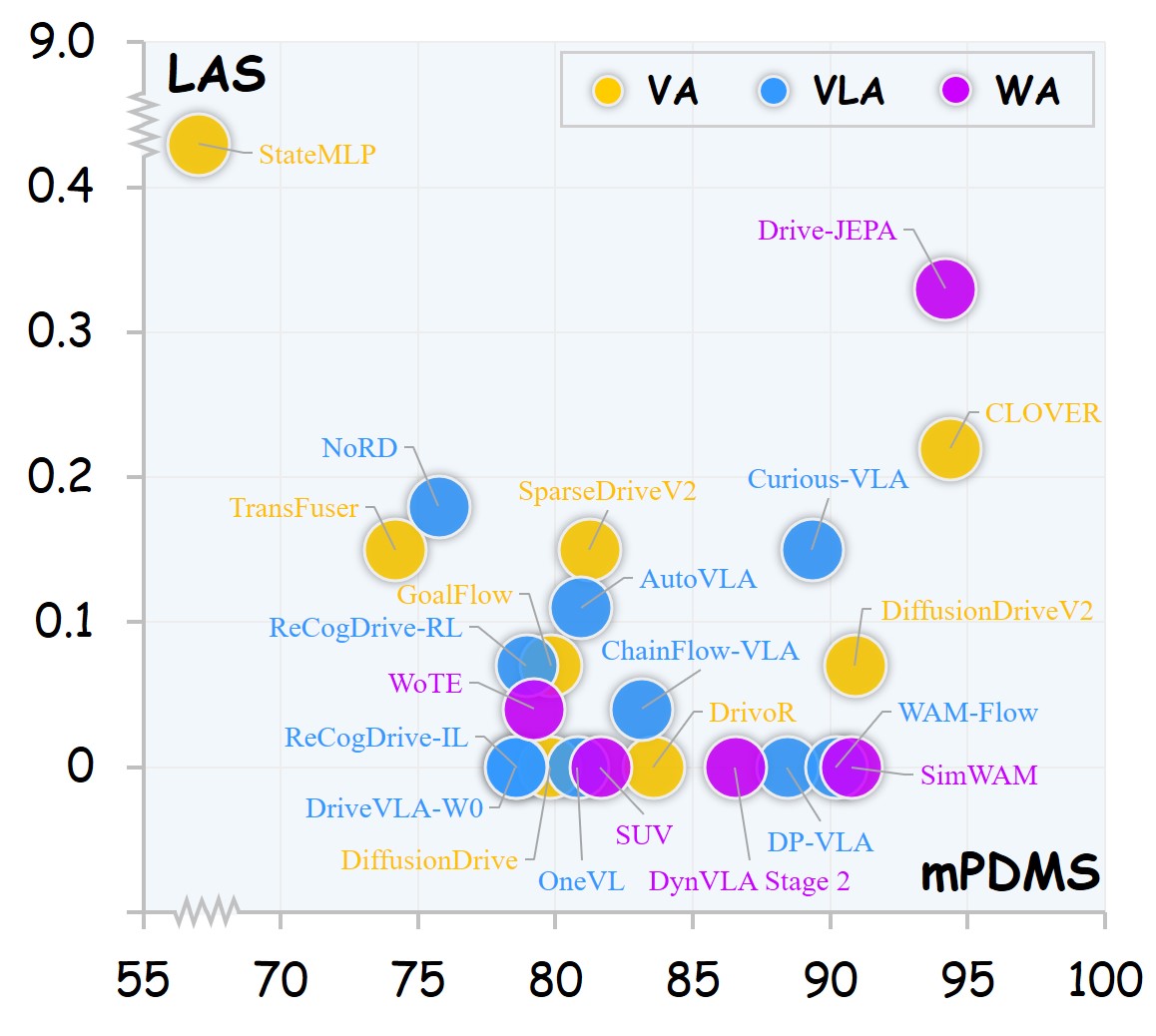}
    \caption{mPDMS \textit{vs}. LAS}
    \label{fig:rel_sub4}
  \end{subfigure}\hfill
  \begin{subfigure}[b]{0.32\textwidth}
    \centering
    \includegraphics[width=\linewidth,height=4.05cm,keepaspectratio]{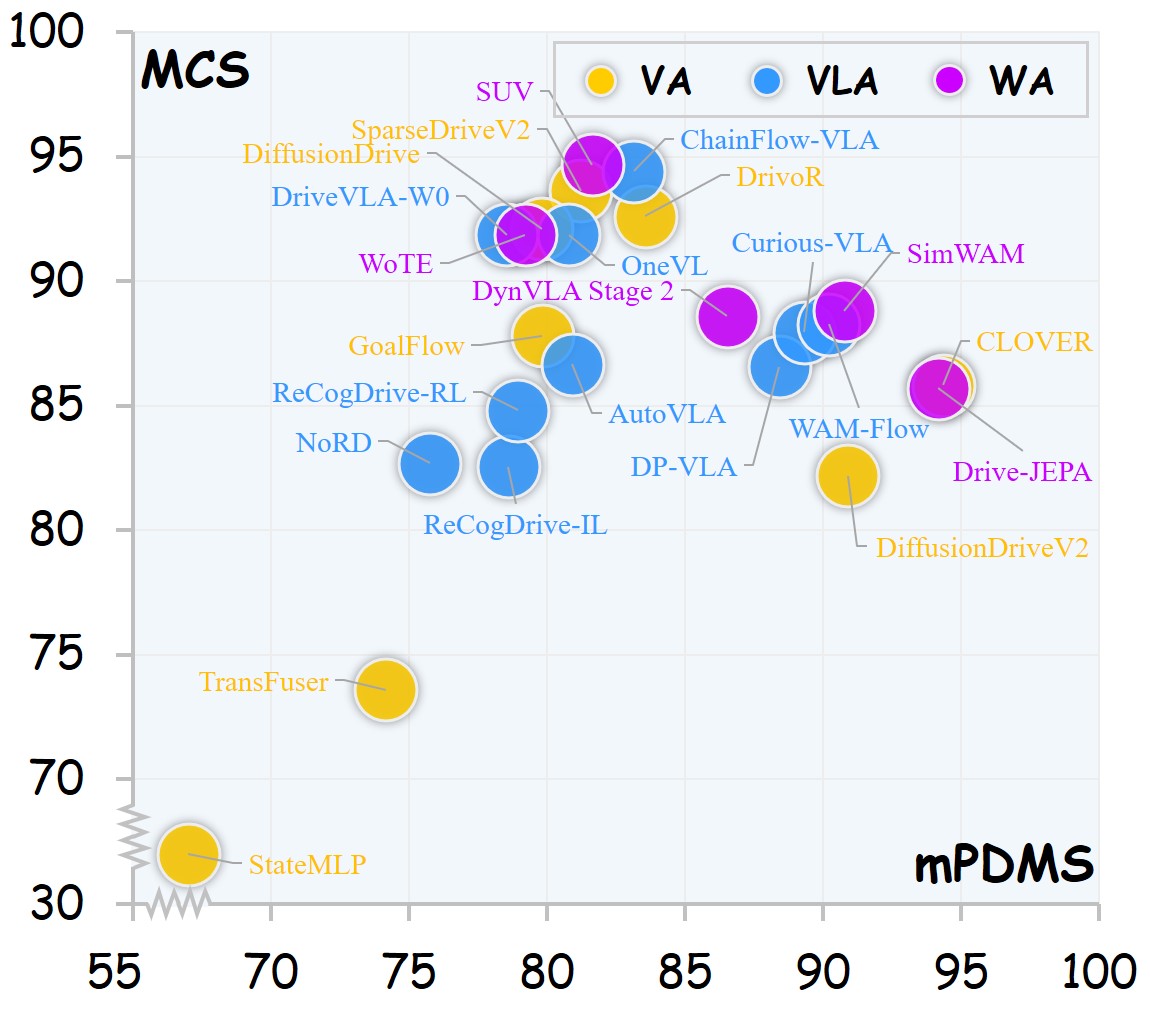}
    \caption{mPDMS \textit{vs}. MCS}
    \label{fig:rel_sub5}
  \end{subfigure}\hfill
  \begin{subfigure}[b]{0.32\textwidth}
    \centering
    \includegraphics[width=\linewidth,height=4.05cm,keepaspectratio]{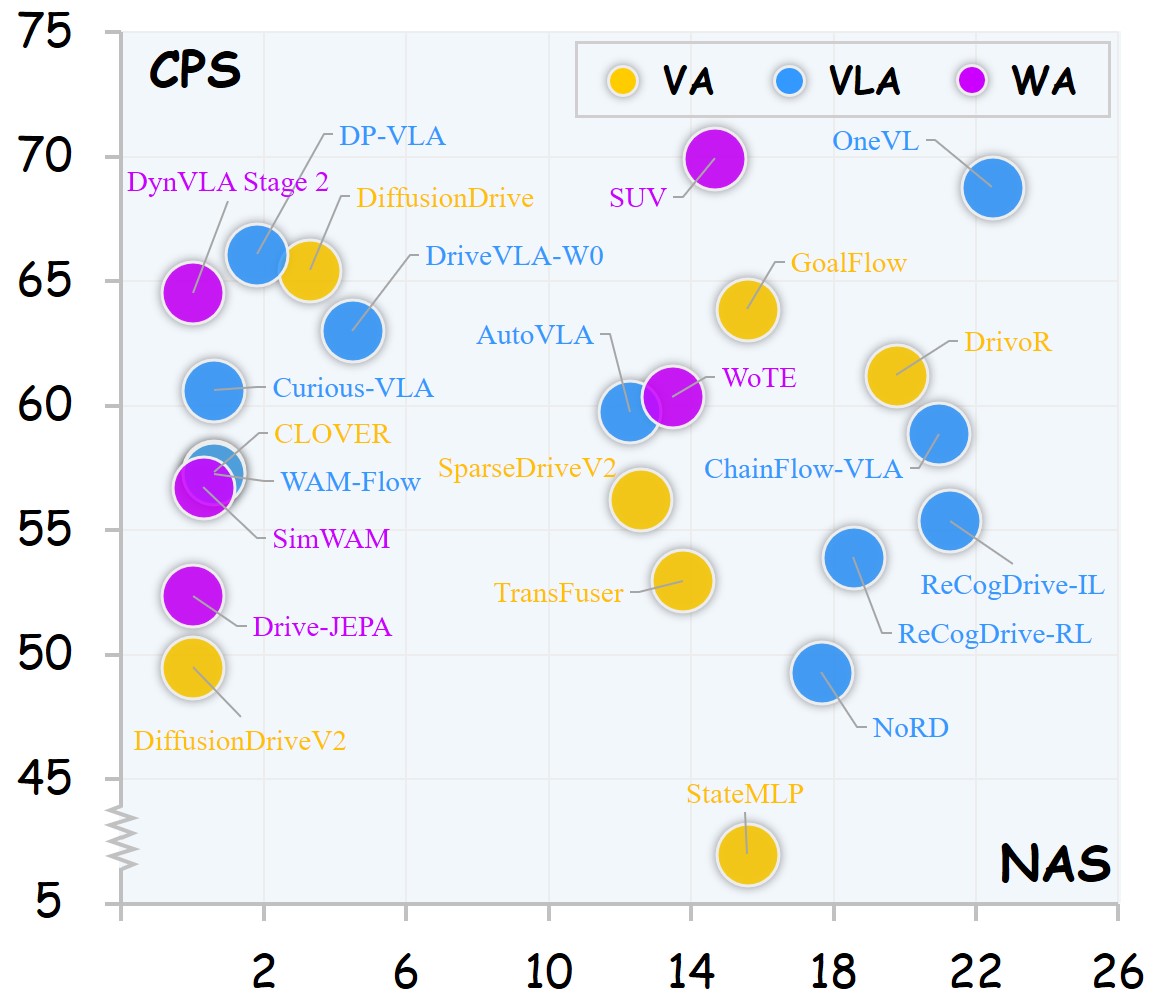}
    \caption{NAS \textit{vs}. CPS}
    \label{fig:nas_cps}
  \end{subfigure}
  \vspace{-2mm}
  \caption{Driving quality and in-context performance across 23 end-to-end driving models.}
  \label{fig:rel_images}
\end{figure}

\subsection{Overall Evaluation}
\label{subsec:4-2-overall_results}

Table~\ref{tab:main_results} presents the overall results of the 23 evaluated models. Figures~\ref{fig:rel_sub1}--\ref{fig:rel_sub5} compare conventional driving quality with route-level and lane-level in-context performance. mPDMS evaluates driving quality under the original navigation condition, while NCS evaluates in-context driving decision-making through paired navigation interventions. The Spearman rank correlation between mPDMS and NCS is $-0.542$, indicating substantial differences in model rankings across the two evaluations. Confidence intervals, pairwise comparisons, and sensitivity analyses of model rankings are provided in Appendices~\ref{sec:appendix_statistical_analysis} and~\ref{sec:appendix_metric_robustness}.

Figure~\ref{fig:rel_images} further reveals two interesting patterns:
(1) High mPDMS does not consistently correspond to high NAS. All models with mPDMS above 85 achieve NAS of at most 1.80, while OneVL achieves the highest NAS of 22.46 with an mPDMS of 80.79.
(2) Different model architecture paradigms do not exhibit a consistent performance trend on \ours. VA, VLA, and WA models all show substantial performance variation within each paradigm, suggesting that the architectural paradigm alone does not consistently explain in-context driving decision-making performance.

\subsection{Model Capability Analysis}
\label{subsec:4-3-capability_analysis}

\begin{wraptable}{r}{0.44\textwidth}
\vspace{-2mm}
\centering
\caption{Contextual success across \ours\ categories.}
\label{tab:category_results}
\vspace{-2mm}
\small
\setlength{\tabcolsep}{7pt}
\renewcommand{\arraystretch}{1.06}
\begin{tabular}{lrr}
\toprule
Category & Instances & Success (\%) \\
\midrule
A   & 334   & 10.01 \\
B1  & 251   & 62.91 \\
B2  & 5,285 & 73.28 \\
C   & 580   & 34.98 \\
L   & 2,716 & 0.45 \\
LC  & 395   & 85.51 \\
\bottomrule
\end{tabular}
\vspace{2pt}
\parbox{\linewidth}{\raggedright\scriptsize Success is averaged over all 23 evaluated models.}
\vspace{-8mm}
\end{wraptable}

Table~\ref{tab:category_results} shows substantial variation in contextual success across the six categories. At the route level, the mean NAS is 10.01, compared with a mean CPS of 57.06. NAS is lower than CPS for 22 of the 23 models, indicating that current models achieve limited success when alternative routes are immediately actionable.

Figure~\ref{fig:nas_cps} further illustrates differences in model performance across NAS and CPS. DynVLA Stage 2 achieves a CPS of 64.54 but has zero NAS. StateMLP achieves a NAS of 15.57 and a CPS of 8.68.

At the lane level, Figures~\ref{fig:rel_sub4} and~\ref{fig:rel_sub5} show limited success in executing feasible alternative lane intentions alongside substantially higher success in ongoing-maneuver continuity. LAS is below 0.34 for 22 of the 23 models, including ten with zero scores. StateMLP achieves the highest LAS of 8.73. The consistently low LAS across models identifies the execution of feasible alternative lane intentions as a common failure mode in lane-level in-context driving decision-making.

\subsection{Context Factor Analysis}
\label{subsec:4-4-context_factor_analysis}

\begin{table*}[htbp]
\centering
\caption{Associations between scene context factors and \ours\ performance.}
\label{tab:scene_context_factors}

\vspace{-2mm}

\small
\setlength{\tabcolsep}{5.5pt}
\renewcommand{\arraystretch}{1.08}

\begin{tabular}{c|ccc|ccc|c}
\toprule
& \multicolumn{3}{c|}{\textbf{Decision Urgency}}
& \multicolumn{3}{c|}{\textbf{Geometric Salience}}
& \textbf{Topological Evidence} \\
\cmidrule(lr){2-4}
\cmidrule(lr){5-7}
Category
& Low & Medium & High
& Low & Medium & High
& $\Delta q$ [95\% CI] \\
\midrule

A
& 15.59 & 11.28 & 3.22
& 6.66 & 12.34 & 11.02
& $+13.15^{*}$ [10.62, 15.72] \\

B1
& 64.22 & 61.90 & 62.63
& 55.74 & 77.43 & 55.49
& $+35.67^{*}$ [21.22, 49.46] \\

B2
& 73.50 & 75.06 & 71.29
& 73.10 & 75.36 & 71.39
& $-13.45^{*}$ [-15.16, -11.78] \\

C
& 23.74 & 43.39 & 37.79
& 29.96 & 36.09 & 38.86
& $-20.77^{*}$ [-28.05, -13.39] \\

L
& 0.49 & 0.43 & 0.42
& 0.46 & 0.43 & 0.46
& $+0.08$ [-0.07, 0.22] \\

LC
& 87.72 & 84.21 & 84.96
& 88.33 & 84.97 & 83.61
& $-2.51$ [-7.62, 3.13] \\

\bottomrule
\end{tabular}

\vspace{2pt}
\begin{minipage}{0.98\textwidth}
\scriptsize
\textbullet\; Decision Urgency and Geometric Salience report contextual success (\%) under within-category Low/Medium/High stratification.\\[-1pt]
\textbullet\; Topological Evidence reports Strong-minus-Weak contextual success. \\[-1pt]
\textbullet\; $^{*}$ indicates that the 95\% scene-token bootstrap confidence interval excludes zero. \\[-1pt]
\textbullet\; Detailed evidence definitions are provided in the Appendix.
\end{minipage}

\end{table*}

We further investigate how in-context driving performance varies with environmental context under different decision requirements. All analyses use the contextual success indicator $q_i$ defined in Section~\ref{subsec:3-3-contextual_evaluation} and aggregate predictions from all 23 evaluated models. All comparisons are performed within the same \ours\ category. We group contextual factors into scene-level factors, including decision timing, road geometry, and topological evidence, and dynamic factors describing traffic interaction.

\paragraph{Scene context.}
Table~\ref{tab:scene_context_factors} reveals substantially different performance patterns across \ours\ categories. For category A, contextual success decreases from 15.59\% under low decision urgency to 3.22\% under high urgency. B1, B2, L, and LC exhibit smaller variations across urgency levels. Geometric salience exhibits no consistent monotonic trend across categories.~B1 reaches its highest performance at the medium level, C shows higher performance at greater salience, whereas LC shows the opposite tendency. Topological evidence provides a clearer distinction. Stronger target-branch directional evidence is associated with 13.15 and 35.67 percentage-point higher contextual success for A and B1, respectively, whereas B2 and C show 13.45 and 20.77 percentage-point lower success under their corresponding strong structural cues. These results show that road context provides different forms of decision evidence depending on the behavioral requirement.

\begin{table*}[htbp]
\centering
\caption{Associations between traffic interaction factors and \ours\ performance.}
\label{tab:traffic_context_factors}

\vspace{-2mm}

\small
\setlength{\tabcolsep}{6pt}
\renewcommand{\arraystretch}{1.08}

\begin{tabular}{c|rrrr|r}
\toprule
Category
& Lead Vehicle
& TTC Risk
& Target Occupancy
& Crossing Conflict
& Traffic Density \\
\midrule

A
& $-4.88$
& $-9.08^{*}$
& $-9.94^{*}$
& $-0.18$
& $+5.31^{*}$ \\

B1
& $+17.47$
& $-5.15$
& $-6.00$
& $+12.39$
& $-0.88$ \\

B2
& $+9.05^{*}$
& $+6.29^{*}$
& $+5.54^{*}$
& $-4.45^{*}$
& $+12.10^{*}$ \\

C
& $+3.58$
& $-10.36$
& $-10.02^{*}$
& $-11.51^{*}$
& $-9.75^{*}$ \\

L
& $-0.50^{*}$
& $-0.26^{*}$
& $-0.39^{*}$
& $+0.20$
& $-0.13^{*}$ \\

LC
& $-7.62^{*}$
& $-8.62^{*}$
& $-6.88^{*}$
& $-9.56^{*}$
& $-7.82^{*}$ \\

\bottomrule
\end{tabular}

\vspace{2pt}
\begin{minipage}{0.98\textwidth}
\scriptsize
\textbullet\; Values report changes in contextual success (pp), computed as comparison minus reference.\\[-1pt]
\textbullet\; $^{*}$ indicates that the 95\% scene-token bootstrap confidence interval excludes zero.
\end{minipage}

\end{table*}

\paragraph{Traffic interaction.}
Table~\ref{tab:traffic_context_factors} shows similarly category-dependent associations between dynamic traffic conditions and contextual success. For Lead Vehicle and TTC, we compare \(\leq 10\) m with \(>25\) m and \(<3\) s with \(\geq 6\) s, respectively. Target Occupancy and Crossing Conflict compare \(\geq 2\) actors with 0 actors, while Traffic Density compares High with Low. A shows the largest performance differences across TTC risk and target-side occupancy, while C shows lower contextual success under greater target-side occupancy and crossing conflicts. LC exhibits consistent performance degradation across all five traffic factors, with lower success under more demanding dynamic conditions. For B2, closer lead vehicles, higher TTC risk, and greater target-side occupancy are associated with higher contextual success. Success increases by 12.10 percentage points from low- to high-density traffic, while crossing conflicts are associated with a decrease of 4.45 percentage points. These associations differ from those observed in LC, where contextual success decreases across all five traffic factors.

\section{Conclusion}
\label{sec:5-Conclusion}

We present \ours, a benchmark for evaluating in-context driving decision-making through paired navigation interventions under fixed observations and vehicle history. Across 9,561 scene--condition instances and 23 end-to-end driving models, we find that strong conventional driving quality does not consistently correspond to strong in-context decision-making. Our analyses further show that model performance varies across different decision requirements and environmental contexts. These results indicate that in-context driving decision-making captures a capability dimension not fully reflected by conventional driving-quality metrics. By evaluating intention--context--decision consistency across route-level and lane-level scenarios, \ours\ provides a structured benchmark for assessing how end-to-end driving models respond to high-level intentions under different driving contexts.

\clearpage
\subsection*{AI Use Statement}

Generative AI tools were used to assist with manuscript drafting and revision, translation, and consistency checks of reported numerical results.  The authors reviewed the AI-assisted content, checked factual claims against the underlying experimental results and cited sources, and made the final decisions on the manuscript. The authors take responsibility for the accuracy and integrity of all content in this work.

\subsection*{Ethics Statement}

\ours\ evaluates in-context driving decision-making using existing driving scenes from NAVSIM \citep{dauner2024navsim} and publicly released or official model assets. All experiments are conducted offline, without interventions on real vehicles or road users. The human validation procedure involves expert inspection of existing scenes and model predictions, as described in Appendix~\ref{sec:appendix_human_validation}. Responsible reuse of the benchmark requires adherence to the source data licenses and protection of privacy in driving-scene imagery.

\subsection*{Reproducibility Statement}

The paired navigation intervention protocol and contextual evaluation framework are specified in Section~\ref{sec:3-PlanContext}. Appendices~\ref{sec:appendix_scenario}, \ref{sec:appendix_evaluation}, and~\ref{sec:appendix_metric} detail scenario construction and filtering, category-specific evaluation rules, geometric thresholds, safety criteria, and metric aggregation. The experimental setup and model evaluation protocol are described in Section~\ref{subsec:4-1-experimental_setup} and Appendix~\ref{sec:appendix_model_evaluation}. The scene-level bootstrap procedure and metric robustness analyses are documented in Appendices~\ref{sec:appendix_statistical_analysis} and~\ref{sec:appendix_metric_robustness}. These specifications support consistent reconstruction of the benchmark and its evaluation procedure. We plan to publicly release the PlanContext benchmark toolkit, including evaluation code and supporting documentation.



\bibliography{iclr2027_conference}
\bibliographystyle{iclr2027_conference}

\clearpage

\appendix
{\LARGE  Appendix}

In the appendix, we provide additional details on the construction, evaluation, and analysis of \ours\ that are omitted from the main paper for brevity. The appendix is organized as follows:
\begin{itemize}
    \item Sec.~\ref{sec:appendix_scenario} provides detailed scenario mining, filtering, and category assignment criteria for the six \ours\ contexts.
    \item Sec.~\ref{sec:appendix_evaluation} presents the category-specific behavioral evaluation rules, including road-relative trajectory comparison, path-change detection, GT-relative suppression, and continuity evaluation for ongoing maneuvers.
    \item Sec.~\ref{sec:appendix_metric} describes road compliance, dynamic safety, and the construction of the final \ours\ metrics.
    \item Sec.~\ref{sec:appendix_experiments} provides additional experimental details, statistical analyses, implementation information, and supplementary results.
    \item Sec.~\ref{sec:appendix_qualitative} presents qualitative examples across contextual categories and additional cases for ten evaluated models.
\end{itemize}

\section{Scenario Construction Details}
\label{sec:appendix_scenario}

This section provides the detailed scenario mining, filtering, and category assignment criteria used to construct \ours. The construction process first removes scenes with insufficient geometric or motion evidence. It then identifies reliable route branches and adjacent lane structures from the local map. Each retained scene--condition instance is assigned to one of the six contextual categories according to a fixed priority order. This procedure is designed to ensure that each category corresponds to a sufficiently clear contextual requirement.

\subsection{Common Validity Filtering}
\label{sec:appendix_common_filtering}

Each candidate instance contains an original navigation intention and an alternative navigation intention selected from \textit{left}, \textit{straight}, and \textit{right}. We retain only valid intention pairs for which the initial road, local map topology, logged trajectory, and the category-required reference geometry and permitted road regions can be reliably reconstructed. A target reference path is required only for categories where a valid target branch or adjacent lane is expected to exist.

We additionally remove near-stationary scenes using the future motion of the logged ego trajectory. Let
\begin{equation}
D_{\mathrm{GT}}^{4s}
=
\sum_t
\left\|
p_t^{\mathrm{GT}}-p_{t-1}^{\mathrm{GT}}
\right\|_2
\end{equation}
denote the cumulative displacement of the logged trajectory over the future four-second horizon. A candidate is retained only when
\begin{equation}
D_{\mathrm{GT}}^{4s} \geq 2.0~\mathrm{m}.
\end{equation}

This motion filtering removes 598 near-stationary candidates from the initial 12,840 candidates and leaves 12,242 candidates for subsequent contextual screening. Scenes in which the road structure, route branch, adjacent lane, or maneuver stage cannot be determined reliably are marked as ambiguous, transition, or pending and are excluded from the main benchmark.

\subsection{Route Branch Construction and Reliability}
\label{sec:appendix_route_construction}

For route-level categories, we search forward from the current road for the first decision point with multiple outgoing edges. The forward search range is limited to 200~m. Each candidate branch is then extended for up to 100~m after the decision point.

Candidate branches are classified according to their terminal heading relative to the reference heading at the decision point. Let $\Delta\theta$ denote this relative heading difference. A branch is classified as straight when
\begin{equation}
|\Delta\theta| \leq 20^\circ.
\end{equation}
It is classified as left when
\begin{equation}
\Delta\theta \geq 30^\circ,
\end{equation}
and as right when
\begin{equation}
\Delta\theta \leq -30^\circ.
\end{equation}
Branches falling between these angular regions are considered ambiguous and are not used for reliable category assignment.

The original route branch is selected primarily according to route overlap. When multiple branches have identical route overlap, the logged trajectory is used only as a secondary geometric cue. The closest branch is selected only when its mean distance to the logged trajectory is at least 0.75~m smaller than that of the second closest branch. Otherwise, the instance is marked as having weak GT-based tie-breaking evidence and is excluded.

The original and alternative paths are sampled at intervals of 0.5~m. Reliable separation is searched only within their common valid path extent after the decision point, with the maximum search range defined as
\begin{equation}
s_{\max}
=
\min(L_N,L_F,50~\mathrm{m}),
\end{equation}
where $L_N$ and $L_F$ denote the valid lengths of the original and alternative paths.

Within this range, the two paths are considered stably separated when three consecutive sampled positions satisfy at least one of the following conditions:
\begin{equation}
\left\|
p^{N}(s)-p^{F}(s)
\right\|_2
\geq
1.5~\mathrm{m},
\end{equation}
or
\begin{equation}
|\Delta\theta(s)|
\geq
12^\circ.
\end{equation}
The earliest position at which this condition is continuously satisfied is recorded as the reliable separation position for the corresponding route pair.

A requested navigation intention may correspond to multiple alternative branches. In this case, the route evidence is considered reliable only when every alternative branch associated with the requested intention exhibits stable separation from the selected original branch. For categories A and B1, a valid arrival time must also be available for every alternative branch. If any branch lacks stable separation or a valid arrival estimate, the robust route evidence is treated as unavailable and the instance cannot be assigned to A or B1.

For an instance with multiple valid alternative branches, we use the latest reliable separation distance and the latest estimated arrival time:
\begin{equation}
d_{\mathrm{split}}^{\mathrm{robust}}
=
\max_j d_{\mathrm{split},j},
\end{equation}
\begin{equation}
t_{\mathrm{arrival}}^{\mathrm{robust}}
=
\max_j t_{\mathrm{arrival},j}.
\end{equation}
Here, $j$ indexes all alternative branches associated with the requested navigation intention. The arrival time of each branch is estimated from the cumulative future distance of the logged trajectory. Linear interpolation is applied between neighboring trajectory samples, which are spaced at 0.5~s intervals.

\subsection{Category Assignment Priority}
\label{sec:appendix_category_priority}

Some scenes satisfy more than one preliminary geometric condition. We therefore apply a fixed category assignment order to avoid overlapping definitions.

A scene is first assigned to category C when strict post-commitment evidence is detected. A scene that starts on a lane connector is excluded when strict commitment evidence is unavailable. If a reliable route divergence can be reached within 2.0~s, the instance is assigned to category A. Cases with an arrival time between 2.0~s and 4.5~s are treated as transition cases and are excluded.

If an active lateral maneuver is already present, the instance is assigned to category LC. Among the remaining cases, a reliable route divergence with an arrival time between 4.5~s and 8.0~s forms a category B1 candidate. A scene is assigned to category L when the ego vehicle remains stably within its current lane and a geometrically feasible adjacent target lane is available. Category B2 is used when no reliable nearby route branch and no feasible adjacent target lane support the alternative intention. All remaining cases are marked as ambiguous or pending and are excluded from the main benchmark.

\subsection{Category-Specific Construction Criteria}
\label{sec:appendix_category_criteria}

\paragraph{Category A: Actionable Route Change.}
Category A contains scenes in which the alternative route has entered the current decision window and should produce a target-directed response within the four-second prediction horizon. A reliable alternative route branch must satisfy
\begin{equation}
0 <
t_{\mathrm{arrival}}^{\mathrm{robust}}
\leq
2.0~\mathrm{s}.
\end{equation}

The ego vehicle must not have strictly committed to the original route. Cases beginning on a lane connector are also excluded when sufficient commitment evidence is unavailable. The interval
\begin{equation}
2.0 <
t_{\mathrm{arrival}}^{\mathrm{robust}}
<
4.5~\mathrm{s}
\end{equation}
is treated as a transition region and does not contribute to either A or B1.

\paragraph{Category B1: Delayed Route Change.}
Category B1 represents scenes in which the alternative route is feasible in the near future but should not yet induce a directional commitment within the current prediction horizon. The reliable arrival time must satisfy
\begin{equation}
4.5
\leq
t_{\mathrm{arrival}}^{\mathrm{robust}}
\leq
8.0~\mathrm{s}.
\end{equation}

We additionally require sufficient remaining distance between the reliable branch separation and the future motion covered by the logged trajectory:
\begin{equation}
d_{\mathrm{split}}^{\mathrm{robust}}
-
D_{\mathrm{GT}}^{4s}
\geq
8.0~\mathrm{m}.
\end{equation}

The scene must not contain a feasible adjacent lane that could already provide a valid immediate interpretation of the alternative navigation intention.

\paragraph{Category B2: Locally Invalid Target.}
Category B2 contains alternative left or right intentions that do not correspond to a locally reachable target in the requested direction. The scene must contain neither a reliable nearby route branch in the requested direction nor a geometrically feasible adjacent target lane within the same roadblock.

We conservatively exclude candidates for which the local map still provides evidence of a potentially reachable target. In particular, a candidate is excluded from B2 whenever at least one branch associated with the alternative direction is detected and its decision point lies within the four-second logged travel distance,
\begin{equation}
N_{\mathrm{replacement}} > 0
\end{equation}
and
\begin{equation}
d_{\mathrm{decision}}
\leq
D_{\mathrm{GT}}^{4s}.
\end{equation}
This exclusion is applied even when the complete route-pair reliability test is not satisfied. Such cases are excluded from B2 and retained as boundary cases for analysis.

Candidates are also excluded when an adjacent lane exists but fails the required geometric continuity checks. These exclusions reduce ambiguity in the B2 set by removing cases with residual evidence of a potentially reachable target.

\paragraph{Category C: Post-Commitment Route Change.}
Category C represents scenes in which the alternative route was available at an earlier decision point and the ego vehicle has already passed the relevant divergence and committed to the original route.

Candidate detection searches backward for up to 80~m to locate the nearest previous decision point with multiple outgoing branches. The current branch must agree with the original left or right navigation intention. The alternative direction must also have existed at that historical decision point. The original and alternative branches must exhibit reliable geometric separation. In addition, the ego heading change over approximately the previous two seconds must be consistent with the original turning direction and must reach at least $10^\circ$.

We measure the post-commitment progress along the selected committed branch as
\begin{equation}
d_{\mathrm{commit}}
=
s_{\mathrm{ego}}^{\mathrm{committed}}
-
s_{\mathrm{divergence}},
\end{equation}
where $s_{\mathrm{ego}}^{\mathrm{committed}}$ denotes the ego progress measured from the beginning of the selected committed branch and $s_{\mathrm{divergence}}$ denotes the stable separation position measured along the same branch.

Candidate detection requires
\begin{equation}
d_{\mathrm{commit}}
\geq
3.0~\mathrm{m}.
\end{equation}

The main benchmark uses the stricter criterion
\begin{equation}
d_{\mathrm{commit}}
\geq
5.0~\mathrm{m}.
\end{equation}

Cases with
\begin{equation}
3.0~\mathrm{m}
\leq
d_{\mathrm{commit}}
<
5.0~\mathrm{m}
\end{equation}
are excluded from the main benchmark and retained as boundary cases for analysis.

\paragraph{Category L: Feasible Adjacent Lane Change.}
Category L contains scenes in which the ego vehicle has not started a lateral maneuver and the adjacent target lane in the requested direction is currently reachable. The ego vehicle must be within 1.0~m of the centerline of the current lane. The nearest lane associated with each of the most recent three history samples must also remain the current lane.

For the target adjacent lane, the relative heading, lateral distance, and parallel overlap must satisfy
\begin{equation}
|\Delta\theta_{\mathrm{lane}}|
\leq
15^\circ,
\end{equation}
\begin{equation}
d_{\mathrm{lateral}}
\leq
8.0~\mathrm{m},
\end{equation}
and
\begin{equation}
L_{\mathrm{parallel}}
\geq
20.0~\mathrm{m}.
\end{equation}

The available map interface does not provide complete legal attributes for every lane boundary. We therefore interpret these cases as topologically feasible adjacent lane changes whose boundary legality cannot always be independently verified. Cases with discontinuous geometry or unstable current lane assignment are excluded.

\paragraph{Category LC: Ongoing Lateral Maneuver.}
Category LC contains scenes in which a persistent lateral motion is already observable before the navigation intention is changed. We use the ego history from $t=-2.0$~s to $t=0$~s with a sampling interval of 0.5~s.

An active lateral direction is detected through one of two motion patterns. In the first pattern, the historical trajectory starts at least 0.5~m closer to an adjacent lane than to the current lane,
\begin{equation}
d_{\mathrm{adj}}(-2s)+0.5
<
d_{\mathrm{current}}(-2s).
\end{equation}
The trajectory must then move at least 0.75~m farther from that adjacent lane,
\begin{equation}
d_{\mathrm{adj}}(0)
-
d_{\mathrm{adj}}(-2s)
\geq
0.75~\mathrm{m},
\end{equation}
while moving at least 0.75~m closer to the current lane,
\begin{equation}
d_{\mathrm{current}}(-2s)
-
d_{\mathrm{current}}(0)
\geq
0.75~\mathrm{m}.
\end{equation}

In the second pattern, the trajectory moves from the current lane toward an adjacent lane. It must become at least 0.75~m closer to the adjacent lane while moving at least 0.75~m farther from the current lane.

The relation between the alternative navigation intention and the detected active direction is recorded as \textit{continue}, \textit{reverse}, or \textit{cancel}. These relations are used for supplementary analysis and for the continuity evaluation described in Sec.~\ref{sec:appendix_evaluation}. They are not treated as separate benchmark categories.

\subsection{Scenario Filtering Statistics}
\label{sec:appendix_filtering_statistics}

Table~\ref{tab:scenario_filtering} summarizes the main filtering stages used to construct \ours. Starting from 12,840 scene--condition candidates, motion filtering retains 12,242 instances. The subsequent context-aware screening and category-specific validity checks yield the final 9,561 instances used for evaluation.

\begin{table}[t]
\centering
\caption{Scenario filtering statistics for \ours.}
\label{tab:scenario_filtering}
\begin{tabular}{lrl}
\toprule
Stage & Instances & Description \\
\midrule
Initial candidates
& 12,840
& Original scene--condition candidates \\

After motion filtering
& 12,242
& Remove near-stationary instances \\

Final benchmark
& 9,561
& Apply final context-aware filtering \\
\bottomrule
\end{tabular}
\end{table}

The final benchmark contains 334 A, 251 B1, 5,285 B2, 580 C, 2,716 L, and 395 LC instances.

\section{Category-Specific Evaluation Rules}
\label{sec:appendix_evaluation}

This section provides the detailed trajectory comparison and category-specific behavioral evaluation rules used by \ours. Given the original prediction $T_i^N$ and the prediction under the alternative navigation intention $T_i^F$, we first represent both trajectories with respect to the reference geometry required by each evaluation component. We then determine their topological outcomes, identify persistent behavioral changes induced by the navigation intervention, and evaluate whether these changes satisfy the contextual requirement of each category.

\subsection{Notation and Reference Geometries}
\label{sec:appendix_reference_geometry}

Each prediction contains eight future ego states sampled at 0.5~s intervals over a four-second horizon. Predicted poses are interpreted at the rear axle center and transformed into the global SE(2) coordinate frame.

We use two reference geometries for different evaluation purposes. The original-route reference is used to characterize the path-level difference between the original and alternative predictions. The logged GT trajectory provides a local reference for the suppression tests of B1, B2, and C and for the continuity evaluation of LC. The GT reference represents the natural road evolution observed in the scene and is not treated as a trajectory target to be imitated.

For the common path-change analysis, each prediction is projected onto the original-route reference. For a projected state, we record the longitudinal road progress $s$, signed lateral offset $d$, and relative heading
\begin{equation}
\theta_r
=
\operatorname{wrap}
\left(
\theta_{\mathrm{ego}}
-
\theta_{\mathrm{road}}
\right).
\end{equation}

The original and alternative predictions are aligned at common road-progress values. We define
\begin{equation}
\Delta d(s)
=
d^F(s)-d^N(s)
\end{equation}
and
\begin{equation}
\Delta\theta(s)
=
\operatorname{wrap}
\left(
\theta_r^F(s)-\theta_r^N(s)
\right).
\end{equation}

This progress-based alignment reduces the influence of longitudinal speed differences when comparing behavioral changes induced by the navigation intervention.

The original-route and target-route references are also used to characterize the terminal topology of each prediction. We denote them as $\mathcal{R}_o$ and $\mathcal{R}_t$. The GT reference used for suppression and continuity evaluation is denoted as $\mathcal{R}_{GT}$.

\subsection{Evaluation Structure}
\label{sec:appendix_evaluation_structure}

The category-specific criteria are constructed from a shared set of trajectory-level primitives, including route-topology assignment, road-relative trajectory change, GT-relative residual change, full-vehicle road compliance, and dynamic safety. Each contextual category selects the evidence required by its behavioral objective.

Categories A and L evaluate whether a feasible alternative intention is executed. Categories B1, B2, and C evaluate whether the alternative navigation condition induces an unnecessary persistent deviation from the locally appropriate behavior. Category LC evaluates the continuity of an ongoing lateral maneuver.

This factorized design keeps the underlying geometric measurements shared across categories while allowing the decision criterion to reflect the corresponding contextual requirement. Continuous trajectory evidence is evaluated using fixed decision thresholds and boundary regions. Clear evidence contributes to the category-specific decision outcome, while boundary-level evidence is retained for uncertainty-aware evaluation. All category definitions and thresholds are fixed and shared across models.

\subsection{Topological Outcome Assignment}
\label{sec:appendix_topology_assignment}

We determine the topological outcome of each predicted trajectory from its terminal portion. Let $d_o$ and $d_t$ denote the distances from the trajectory tail to the original-route reference $\mathcal{R}_o$ and target-route reference $\mathcal{R}_t$.

When both reference distances are available, a trajectory is assigned to the target topology when
\begin{equation}
d_t+0.5 < d_o.
\end{equation}

When
\begin{equation}
d_o+0.5 < d_t,
\end{equation}
the trajectory is assigned to the original topology for categories A and L. The same geometric condition is labeled as committed for category C and as active maneuver for category LC.

When the difference between $d_o$ and $d_t$ is smaller than 0.5~m, category B1 uses the shared label. The corresponding outcome is marked as ambiguous for the remaining categories.

Reference geometry can be incomplete for some local configurations. When only $d_o$ is available, the prediction is assigned to the original topology for A, B1, B2, and L. The corresponding labels are committed for C and active maneuver for LC.

When $d_o$ is unavailable, category B2 uses an auxiliary state denoted as \textit{other-local}. The remaining categories use the ambiguous state. The \textit{other-local} state indicates insufficient comparative evidence to assign the prediction to either the original topology or a supported target topology. It does not indicate road compliance or traffic-rule legality. Road compliance is evaluated independently using the full vehicle footprint in Sec.~\ref{sec:appendix_metric}.

The resulting topology variable is denoted by $\tau_i^N$ or $\tau_i^F$ for the original and alternative predictions.

\subsection{Persistent Path-Change Detection}
\label{sec:appendix_path_change}

We next determine whether changing the navigation intention produces a persistent change in the predicted path. Let the trajectory tail contain the final three valid aligned samples. We compute
\begin{equation}
\widetilde{d}_{\mathrm{tail}}
=
\operatorname{median}
\left(
\Delta d_{\mathrm{tail}}
\right),
\end{equation}
\begin{equation}
d_{\mathrm{end}}
=
\Delta d(s_{\mathrm{end}}),
\end{equation}
\begin{equation}
d_{75}
=
Q_{0.75}
\left(
|\Delta d|
\right),
\end{equation}
and
\begin{equation}
\widetilde{\theta}_{\mathrm{tail}}
=
\operatorname{median}
\left(
\Delta\theta_{\mathrm{tail}}
\right).
\end{equation}

For samples satisfying $|\Delta d|\geq0.15$~m, we compute the sign-consistency ratio $r_{\mathrm{sign}}$. This quantity measures the fraction of valid samples whose lateral-change sign agrees with the dominant direction of the trajectory tail.

A strong path change is identified when at least one of the following conditions is satisfied.

\begin{enumerate}
    \item The original and alternative predictions are assigned to different road topologies.

    \item At least four aligned samples are available and
    \begin{equation}
    |\widetilde{d}_{\mathrm{tail}}|
    \geq
    0.85~\mathrm{m},
    \end{equation}
    \begin{equation}
    |d_{\mathrm{end}}|
    \geq
    0.85~\mathrm{m},
    \end{equation}
    \begin{equation}
    d_{75}
    \geq
    0.60~\mathrm{m},
    \end{equation}
    and
    \begin{equation}
    r_{\mathrm{sign}}
    \geq
    0.80.
    \end{equation}

    \item The tail heading and lateral differences satisfy
    \begin{equation}
    |\widetilde{\theta}_{\mathrm{tail}}|
    \geq
    8.5^\circ,
    \end{equation}
    \begin{equation}
    |\widetilde{d}_{\mathrm{tail}}|
    \geq
    0.60~\mathrm{m},
    \end{equation}
    and
    \begin{equation}
    r_{\mathrm{sign}}
    \geq
    0.80.
    \end{equation}
\end{enumerate}

We additionally define a boundary-evidence region for diagnostic review. The lateral criterion uses 0.65~m for the tail and endpoint differences, 0.40~m for $d_{75}$, and 0.70 for the sign-consistency ratio. The heading-based criterion uses a heading difference of $6.5^\circ$, a lateral difference of at least 0.40~m, and a sign-consistency ratio of 0.70.

If no strong path change is detected and
\begin{equation}
\left|
s_{\mathrm{end}}^F
-
s_{\mathrm{end}}^N
\right|
\geq
1.0~\mathrm{m},
\end{equation}
the response is labeled as a progress-only change. Responses without a strong path change or sufficient longitudinal difference are labeled as no change. Cases with insufficient common progress or unavailable reference geometry are labeled as geometry ambiguous.

\subsection{Direction of the Path Change}
\label{sec:appendix_path_direction}

For a persistent path change, we next determine whether the change is directed toward the alternative target, toward the original route, or toward an unrelated direction.

We first apply topology-based decisions. A transition from a non-target topology under the original condition to the target topology under the alternative condition is labeled as a topology switch. If the original prediction is assigned to the target topology and the alternative prediction leaves the target topology, the change is labeled as original-directed.

For the remaining persistent path changes, let
\begin{equation}
g(T)
=
d(T,\mathcal{R}_o)
-
d(T,\mathcal{R}_t)
\end{equation}
denote the relative target preference of trajectory $T$. We define
\begin{equation}
\Delta g
=
g(T^F)-g(T^N).
\end{equation}

When the required route distances are available, the change is labeled as target-directed when
\begin{equation}
\Delta g
\geq
1.2~\mathrm{m}.
\end{equation}

The change is labeled as original-directed when
\begin{equation}
\Delta g
\leq
-1.2~\mathrm{m}.
\end{equation}

All remaining persistent directional changes are labeled as unrelated-directional.

\subsection{GT-Relative Suppression for B1, B2, and C}
\label{sec:appendix_gt_relative}

Categories B1, B2, and C require the alternative navigation intention to avoid inducing an unnecessary persistent path deviation. The logged trajectory is used as a reference for the natural local road evolution. It is not treated as the unique correct trajectory under the alternative navigation condition.

Both predictions are projected onto the GT reference $\mathcal{R}_{GT}$. We define the tail lateral residuals as
\begin{equation}
r_d^N
=
\left|
\operatorname{median}
\left(
d_{\mathrm{tail}}^N
\right)
\right|
\end{equation}
and
\begin{equation}
r_d^F
=
\left|
\operatorname{median}
\left(
d_{\mathrm{tail}}^F
\right)
\right|.
\end{equation}

The corresponding heading residuals are
\begin{equation}
r_\theta^N
=
\left|
\operatorname{median}
\left(
\theta_{\mathrm{tail}}^N
\right)
\right|
\end{equation}
and
\begin{equation}
r_\theta^F
=
\left|
\operatorname{median}
\left(
\theta_{\mathrm{tail}}^F
\right)
\right|.
\end{equation}

The additional deviation induced by the alternative navigation condition is
\begin{equation}
\Delta r_d
=
r_d^F-r_d^N
\end{equation}
and
\begin{equation}
\Delta r_\theta
=
r_\theta^F-r_\theta^N.
\end{equation}

An unnecessary persistent response is identified only when a strong path change has already been detected. Strong lateral evidence is present when
\begin{equation}
r_d^F
\geq
1.5~\mathrm{m}
\end{equation}
and
\begin{equation}
\Delta r_d
\geq
1.0~\mathrm{m}.
\end{equation}

Strong heading evidence is present when
\begin{equation}
r_\theta^F
\geq
12^\circ
\end{equation}
and
\begin{equation}
\Delta r_\theta
\geq
10^\circ.
\end{equation}

Two additional conditions provide strong evidence of an unnecessary response. The first occurs when the original and alternative predictions lie on opposite sides of the GT reference and satisfy
\begin{equation}
r_d^F
\geq
0.85~\mathrm{m}
\end{equation}
and
\begin{equation}
r_d^N
\geq
0.40~\mathrm{m}.
\end{equation}

The second occurs when the alternative prediction switches from a non-target topology to the target topology.

We denote the resulting unnecessary-response indicator as $U_i$. The GT-relative evaluation is considered available when the required GT reference geometry and residual quantities can be computed. We denote this availability by $G_i$.

Boundary evidence is retained for diagnostic review. The GT-relative boundary thresholds use a lateral residual of at least 1.0~m with an additional increase of at least 0.5~m, or a heading residual of at least $8^\circ$ with an additional increase of at least $6^\circ$. A sample is also retained for diagnostic review when the common path-change detector in Sec.~\ref{sec:appendix_path_change} produces only boundary evidence. Diagnostic boundary evidence does not automatically make the contextual decision incorrect.

\subsection{Category-Specific Behavioral Correctness}
\label{sec:appendix_category_evaluation}

For categories A, B1, B2, C, and L, contextual correctness combines the semantic validity of the two predictions with their paired behavioral relationship.

\paragraph{Category A.}
The original prediction is semantically valid when
\begin{equation}
S_i^N
=
[\tau_i^N=\mathrm{original}],
\end{equation}
and the alternative prediction is semantically valid when
\begin{equation}
S_i^F
=
[\tau_i^F=\mathrm{target}].
\end{equation}

The paired behavioral relation is correct when the detected path change is target-directed or produces a topology switch:
\begin{equation}
P_i
=
[
\mathrm{target\mbox{-}directed}
\ \mathrm{or}\
\mathrm{topology\ switch}
].
\end{equation}

The contextual decision correctness is
\begin{equation}
C_i
=
S_i^N
\land
S_i^F
\land
P_i.
\end{equation}

A missing response, an incorrect branch transition, or a persistent change in an unrelated direction is counted as incorrect.

\paragraph{Category B1.}
The original and alternative predictions are semantically valid when they remain on the shared approach or original route:
\begin{equation}
S_i^N
=
[
\tau_i^N
\in
\{
\mathrm{shared},
\mathrm{original}
\}
],
\end{equation}
\begin{equation}
S_i^F
=
[
\tau_i^F
\in
\{
\mathrm{shared},
\mathrm{original}
\}
].
\end{equation}

The paired behavioral requirement is
\begin{equation}
P_i
=
G_i
\land
\neg U_i,
\end{equation}
and the contextual decision correctness is
\begin{equation}
C_i
=
S_i^N
\land
S_i^F
\land
P_i.
\end{equation}

Longitudinal slowing, braking, progress-only changes, and natural road curvature remain valid. Any unnecessary persistent path deviation induced by the alternative intention is counted as incorrect. In B1, the most direct failure mode is premature commitment toward a target branch that has not yet entered the current decision window.

\paragraph{Category B2.}
The original and alternative predictions are semantically valid when they do not enter the unsupported target topology:
\begin{equation}
S_i^N
=
[
\tau_i^N
\neq
\mathrm{target}
],
\end{equation}
\begin{equation}
S_i^F
=
[
\tau_i^F
\neq
\mathrm{target}
].
\end{equation}

The paired behavioral requirement is
\begin{equation}
P_i
=
G_i
\land
\neg U_i,
\end{equation}
which gives
\begin{equation}
C_i
=
S_i^N
\land
S_i^F
\land
P_i.
\end{equation}

A lack of directional response is compatible with B2 but does not independently guarantee final success. Any unnecessary persistent path deviation induced by the alternative intention is counted as incorrect. This includes a persistent response toward an unsupported target and a strong unrelated directional deviation. Road compliance is evaluated separately in Sec.~\ref{sec:appendix_metric}.

\paragraph{Category C.}
The original and alternative predictions must remain consistent with the already committed route:
\begin{equation}
S_i^N
=
[
\tau_i^N
\in
\{
\mathrm{committed},
\mathrm{original}
\}
],
\end{equation}
\begin{equation}
S_i^F
=
[
\tau_i^F
\in
\{
\mathrm{committed},
\mathrm{original}
\}
].
\end{equation}

The paired behavioral requirement is
\begin{equation}
P_i
=
G_i
\land
\neg U_i,
\end{equation}
and
\begin{equation}
C_i
=
S_i^N
\land
S_i^F
\land
P_i.
\end{equation}

The model may slow down, stop, or modify its longitudinal progress. Any unnecessary persistent deviation induced by the alternative intention is counted as incorrect. Typical failures include reversal toward a previously missed alternative branch, an additional pull toward that branch, and a strong unrelated directional deviation.

\paragraph{Category L.}
The original prediction is required to remain in the original lane:
\begin{equation}
S_i^N
=
[
\tau_i^N
=
\mathrm{original}
].
\end{equation}

The alternative prediction is required to enter the target adjacent lane:
\begin{equation}
S_i^F
=
[
\tau_i^F
=
\mathrm{target}
].
\end{equation}

The paired behavioral relation is correct when the path change is target-directed or produces a topology switch:
\begin{equation}
P_i
=
[
\mathrm{target\mbox{-}directed}
\ \mathrm{or}\
\mathrm{topology\ switch}
].
\end{equation}

The contextual decision correctness is
\begin{equation}
C_i
=
S_i^N
\land
S_i^F
\land
P_i.
\end{equation}

A small lateral or heading response that remains within the current lane is not counted as a completed lane change.

\subsection{LC Continuity Evaluation}
\label{sec:appendix_lc_evaluation}

Category LC evaluates whether changing the navigation intention disrupts an ongoing lateral maneuver. It does not use independent semantic labels for the original and alternative predictions. The evaluation produces one of three states:
\begin{equation}
P_i^{LC}
\in
\{
\mathrm{correct},
\mathrm{incorrect},
\mathrm{uncertain}
\}.
\end{equation}

Both predictions are projected onto the GT reference $\mathcal{R}_{GT}$. At each future time step $t$, we define the signed lateral residuals
\begin{equation}
d_t^N
=
d(T_t^N,\mathcal{R}_{GT}),
\end{equation}
\begin{equation}
d_t^F
=
d(T_t^F,\mathcal{R}_{GT}),
\end{equation}
and the corresponding heading residuals
\begin{equation}
\theta_t^N
=
\operatorname{wrap}
\left(
\theta(T_t^N)
-
\theta_{GT}(s_t)
\right),
\end{equation}
\begin{equation}
\theta_t^F
=
\operatorname{wrap}
\left(
\theta(T_t^F)
-
\theta_{GT}(s_t)
\right).
\end{equation}

The absolute residual measures deviation from the natural GT road evolution. The difference between the alternative and original residuals is used to distinguish intervention-induced changes from deviations that are already present in the original prediction.

The continuity protection window contains the first three predicted states at 0.5~s, 1.0~s, and 1.5~s.

\paragraph{Early Additional Deviation.}
A strong early lateral deviation is detected when at least two states within the protection window satisfy
\begin{equation}
|d_t^F|
\geq
1.0~\mathrm{m}
\end{equation}
and
\begin{equation}
|d_t^F|
-
|d_t^N|
\geq
0.75~\mathrm{m}.
\end{equation}

A strong early heading deviation is detected when at least two states satisfy
\begin{equation}
|\theta_t^F|
\geq
10^\circ
\end{equation}
and
\begin{equation}
|\theta_t^F|
-
|\theta_t^N|
\geq
8^\circ.
\end{equation}

The corresponding boundary thresholds are 0.75~m and 0.50~m for lateral deviation and $8^\circ$ and $6^\circ$ for heading deviation.

\paragraph{Persistent Reversal Within the Protection Window.}
Using the active maneuver direction detected from the history, we define the opposite lateral direction as the reversal direction. Let $d_{\mathrm{reverse}}^N$ and $d_{\mathrm{reverse}}^F$ denote the projected displacement components along this reversal direction.

A strong persistent reversal is detected when at least two states within the protection window satisfy
\begin{equation}
d_{\mathrm{reverse}}^F
\geq
1.0~\mathrm{m}
\end{equation}
and
\begin{equation}
d_{\mathrm{reverse}}^F
-
d_{\mathrm{reverse}}^N
\geq
0.75~\mathrm{m}.
\end{equation}

The corresponding boundary thresholds are 0.75~m and 0.50~m.

\paragraph{Reversal Beginning Near the End of the Protection Window.}
For the \textit{reverse} relation, we additionally detect sustained reversals that begin near the end of the protection window.

Let $\Delta d_{\mathrm{rev}}(t)$ denote the additional lateral deviation of the alternative prediction relative to the original prediction along the reversal direction at future time $t$. Similarly, $\Delta\theta_{\mathrm{rev}}(t)$ denotes the additional wrapped heading deviation along the reversal direction. Positive values indicate increasing deviation in the reversal direction.

The reversal consistency ratio $r_{\mathrm{rev}}$ is defined as the fraction of valid aligned future states whose signed heading change is consistent with the reversal direction.

Strong evidence requires
\begin{equation}
\Delta\theta_{\mathrm{rev}}(1.5s)
\geq
12^\circ,
\end{equation}
\begin{equation}
\Delta\theta_{\mathrm{rev}}(2.0s)
\geq
18^\circ,
\end{equation}
\begin{equation}
\Delta\theta_{\mathrm{rev}}(2.0s)
-
\Delta\theta_{\mathrm{rev}}(1.5s)
\geq
3^\circ,
\end{equation}
\begin{equation}
\Delta d_{\mathrm{rev}}(2.0s)
\geq
0.75~\mathrm{m},
\end{equation}
\begin{equation}
\operatorname{median}_{t>2s}
\Delta\theta_{\mathrm{rev}}(t)
\geq
20^\circ,
\end{equation}
and
\begin{equation}
r_{\mathrm{rev}}
\geq
0.75.
\end{equation}

The corresponding boundary thresholds are $8^\circ$, $12^\circ$, $2^\circ$, 0.50~m, $15^\circ$, and 0.75.

\paragraph{Unrelated Persistent Deviation.}
An unrelated-directional path change is treated as strong continuity-disruption evidence when the alternative prediction also exhibits a sufficiently large residual relative to the GT reference.

The lateral condition is
\begin{equation}
r_d^F
\geq
1.5~\mathrm{m}
\end{equation}
and
\begin{equation}
\Delta r_d
\geq
1.0~\mathrm{m}.
\end{equation}

The heading condition is
\begin{equation}
r_\theta^F
\geq
12^\circ
\end{equation}
and
\begin{equation}
\Delta r_\theta
\geq
10^\circ.
\end{equation}

An unrelated-directional path change satisfying either condition is counted as strong disruption evidence.

The corresponding boundary thresholds use a lateral residual of at least 1.0~m with an additional increase of at least 0.5~m, or a heading residual of at least $8^\circ$ with an additional increase of at least $6^\circ$.

\paragraph{Final LC Status.}
The LC continuity status is assigned as
\begin{equation}
P_i^{LC}
=
\begin{cases}
\mathrm{incorrect}, & \text{if strong disruption evidence is present}, \\
\mathrm{uncertain}, & \text{if only boundary evidence is present}, \\
\mathrm{correct}, & \text{otherwise}.
\end{cases}
\end{equation}

A smooth response to the new navigation intention after the protection window is allowed. Large absolute deviation from the logged trajectory is retained as diagnostic information and does not independently determine the LC status.

For LC, contextual correctness is
\begin{equation}
C_i
=
1
\end{equation}
when $P_i^{LC}=\mathrm{correct}$ and
\begin{equation}
C_i
=
0
\end{equation}
when $P_i^{LC}=\mathrm{incorrect}$. Instances with $P_i^{LC}=\mathrm{uncertain}$ are excluded from subsequent success-rate aggregation.

\section{Road Compliance, Safety, and Metric Details}
\label{sec:appendix_metric}

This section describes the road-compliance and dynamic-safety evaluation used by \ours, followed by the aggregation of instance-level results into the final benchmark metrics. These components are evaluated separately from the contextual decision correctness defined in Sec.~\ref{sec:appendix_evaluation}.

\subsection{Full-Vehicle Road Compliance}
\label{sec:appendix_road_compliance}

Road compliance is evaluated using the complete ego-vehicle footprint at every predicted future state. Each predicted pose is interpreted at the rear axle center. We use the dimensions of the Pacifica vehicle model, with a length of 5.176~m and a width of 2.297~m. The rear axle is located 1.461~m behind the geometric center of the vehicle footprint.

Let $\mathcal{B}(T_{i,t}^F)$ denote the rectangular vehicle footprint corresponding to the alternative prediction at future state $t$. Let $\Omega_i^F$ denote the road region permitted by the contextual requirement of instance $i$. A 0.35~m geometric buffer is applied to the permitted road region.

The road-compliance indicator is defined as
\begin{equation}
R_i^F
=
\prod_{t=1}^{8}
\mathbb{1}
\left[
\mathcal{B}(T_{i,t}^F)
\subseteq
\Omega_i^F
\right].
\end{equation}

A prediction passes the road-compliance evaluation only when the complete vehicle footprint is contained in the permitted region at all eight future states.

The permitted road region depends on the contextual category. Table~\ref{tab:permitted_road_regions} summarizes the corresponding regions under the original and alternative navigation conditions.

\begin{table}[t]
\centering
\caption{Permitted road regions for the six \ours\ categories.}
\label{tab:permitted_road_regions}
\begin{tabular}{lll}
\toprule
Category & Original condition & Alternative condition \\
\midrule
A  & Original branch & Target branch \\
B1 & Original approach branch & Original approach branch \\
B2 & Local valid roads and adjacent roads & Local valid roads and adjacent roads \\
C  & Committed original branch & Committed original branch \\
L  & Current lane & Current and target adjacent lanes \\
LC & Maneuver-related local roads & Maneuver-related local roads \\
\bottomrule
\end{tabular}
\end{table}

For A, the permitted region follows the original route under the original navigation condition and the target route under the alternative condition. For B1, both conditions remain within the original approach region because the alternative route has not yet entered the current decision window. B2 uses the locally available road region because no supported target route is defined. Category C follows the already committed original route. Category L uses the current lane under the original condition and the union of the current and target adjacent lanes under the alternative condition. Category LC uses the local road region associated with the ongoing lateral maneuver.

Each permitted corridor is extended sufficiently far to cover the longest predicted trajectory arc length, the vehicle front overhang, and an additional 5~m margin. For category A, a target branch that terminates within the local map is allowed to end at its valid road termination.

\subsection{Dynamic and Rule-Based Safety}
\label{sec:appendix_dynamic_safety}

Dynamic safety is evaluated using the NAVSIM safety criteria on the alternative prediction. The evaluation includes collision avoidance, drivable-area compliance, driving-direction compliance, traffic-light compliance, time-to-collision, and history-based comfort.

Let $e_{i,m}^F$ denote the binary result of safety component $m$ for instance $i$. The dynamic-safety indicator is
\begin{equation}
E_i^F
=
\bigwedge_m
e_{i,m}^F.
\end{equation}

The evaluated components are
\begin{equation}
m
\in
\{
\mathrm{collision},
\mathrm{drivable},
\mathrm{direction},
\mathrm{traffic\ light},
\mathrm{TTC},
\mathrm{comfort}
\}.
\end{equation}

The overall safety condition combines the full-vehicle road-compliance check with the NAVSIM safety criteria:
\begin{equation}
\mathrm{OverallSafety}_i
=
R_i^F
\land
E_i^F.
\end{equation}

Road compliance and NAVSIM dynamic safety are kept as separate components because they characterize different failure modes. The former evaluates whether the complete predicted vehicle footprint remains within the context-specific permitted region. The latter evaluates dynamic interactions and rule-related safety conditions.

\subsection{Instance-Level Contextual Success}
\label{sec:appendix_instance_success}

Let $C_i$ denote the contextual decision correctness defined in Sec.~\ref{sec:appendix_evaluation}. The final contextual success of instance $i$ is
\begin{equation}
q_i
=
C_i
\land
R_i^F
\land
E_i^F.
\end{equation}

A successful instance therefore requires a context-consistent decision response, road-compliant motion, and dynamic safety.

For additional analysis, we additionally define the correctness of the alternative-condition prediction as
\begin{equation}
A_i^F
=
\begin{cases}
S_i^F,
& k_i\in\{A,L\}, \\[4pt]
S_i^F
\land
G_i
\land
\neg U_i,
& k_i\in\{B1,B2,C\}, \\[4pt]
\mathbb{1}
\left[
P_i^{LC}
=
\mathrm{correct}
\right],
& k_i=LC.
\end{cases}
\end{equation}

The corresponding alternative-condition success indicator is
\begin{equation}
q_i^{\mathrm{alt}}
=
A_i^F
\land
R_i^F
\land
E_i^F.
\end{equation}

This metric evaluates whether the prediction under the alternative navigation condition is contextually valid, road compliant, and dynamically safe. For B1, B2, and C, alternative-condition validity additionally requires that the GT-relative suppression test is available and that no unnecessary persistent response is detected. For LC, the alternative-condition correctness is determined by the continuity status defined in Sec.~\ref{sec:appendix_lc_evaluation}.

LC instances with \(P_i^{LC}=\mathrm{uncertain}\) are excluded from the valid denominator of the alternative-condition success rate.

\subsection{Class-Balanced Metric Aggregation}
\label{sec:appendix_metric_aggregation}

The six \ours\ categories contain substantially different numbers of instances. We therefore compute the contextual success rate independently within each category before aggregating across categories.

For category $k$, the contextual success rate is
\begin{equation}
S_k
=
\frac{1}{|\mathcal{D}_k|}
\sum_{i\in\mathcal{D}_k}
q_i,
\end{equation}
where $\mathcal{D}_k$ denotes the valid scene--condition instances used for metric aggregation in category $k$. For LC, instances labeled as \textit{uncertain} are excluded from $\mathcal{D}_{LC}$.

We aggregate categories A, B1, B2, and C for route-level evaluation and report categories L and LC separately for lane-level evaluation.

The Navigation Action Score (NAS) is the contextual success rate in category A, where the alternative route is immediately actionable. The Context Preservation Score (CPS) averages the contextual success rates across B1, B2, and C, where the alternative navigation intention should be deferred or rejected:
\begin{equation}
\mathrm{CPS}
=
\frac{
S_{B1}
+
S_{B2}
+
S_C
}{3}.
\end{equation}

The Navigation Context Score (NCS) summarizes route-level in-context driving decision-making through the geometric mean of NAS and CPS:
\begin{equation}
\mathrm{NCS}
=
\sqrt{
\mathrm{NAS}
\cdot
\mathrm{CPS}
}.
\end{equation}

CPS is a macro-average of the success rates in B1, B2, and C, with each category-level score assigned a weight of one third. The geometric mean penalizes imbalanced performance across NAS and CPS.

For lane-level in-context driving decision-making, the Lane Adaptation Score (LAS) and Maneuver Continuity Score (MCS) are the contextual success rates in L and LC, respectively. These scores are reported separately and are not included in NCS.

For comparison with conventional driving quality, matched PDMS is computed on the same 5,911 physical scenes used by \ours\ under the original navigation condition. Each physical scene contributes once to matched PDMS even when it is associated with multiple alternative navigation conditions.

\section{Additional Experimental Details and Analyses}
\label{sec:appendix_experiments}

This section provides additional details on the benchmark composition, model evaluation protocol, statistical analysis, contextual factor definitions, and robustness analyses used in \ours. The definitions of scenario construction, contextual correctness, road compliance, and metric aggregation are provided in Secs.~\ref{sec:appendix_scenario}, \ref{sec:appendix_evaluation}, and \ref{sec:appendix_metric}.

\subsection{Benchmark Composition}
\label{sec:appendix_benchmark_composition}

\ours\ is constructed from the NAVSIM navtest split. The final benchmark contains 9,561 scene--condition instances derived from 5,911 unique physical scenes. A physical scene may contribute multiple scene--condition instances when multiple alternative navigation intentions satisfy the construction criteria.

The six contextual categories contain 334 A, 251 B1, 5,285 B2, 580 C, 2,716 L, and 395 LC instances. Table~\ref{tab:appendix_benchmark_composition} summarizes the benchmark composition.

\begin{table}[t]
\centering
\caption{Composition of the \ours\ benchmark.}
\label{tab:appendix_benchmark_composition}
\begin{tabular}{lrr}
\toprule
Category & Requirement & Instances \\
\midrule
A  & Follow   & 334 \\
B1 & Wait     & 251 \\
B2 & Reject   & 5,285 \\
C  & Reject   & 580 \\
L  & Follow   & 2,716 \\
LC & Preserve & 395 \\
\midrule
Total & & 9,561 \\
\bottomrule
\end{tabular}
\end{table}

The benchmark is category imbalanced because the six contextual conditions occur at different frequencies in NAVSIM. We therefore use the class-balanced aggregation defined in Sec.~\ref{sec:appendix_metric_aggregation} instead of a direct micro-average over all instances.

\subsection{Model Evaluation Protocol}
\label{sec:appendix_model_evaluation}

We evaluate 23 publicly available end-to-end driving models covering three representative paradigms. Vision and state to action models directly predict future trajectories from sensory observations or state inputs. Vision language action models incorporate vision language representations into driving policies. World model based action models explicitly predict or encode future scene dynamics before producing driving actions.

All models are evaluated using publicly released or official assets. Navigation conditions are explicitly involved during inference. Model outputs are standardized to eight future ego poses sampled over a four-second prediction horizon.

For each scene--condition instance, the model is evaluated twice. The first inference uses the original navigation intention $c_i^N$. The second inference uses the alternative navigation intention $c_i^F$. Sensor observations and vehicle history remain unchanged between the two evaluations. This paired protocol isolates the behavioral response to the navigation intervention while preserving the observed scene.

For conventional driving evaluation, we report PDMS on the full NAVSIM navtest split. We additionally compute matched PDMS, denoted as mPDMS, on the same 5,911 physical scenes used by \ours. mPDMS is evaluated under the original navigation condition. Each physical scene contributes once to mPDMS even when it produces multiple scene--condition instances in \ours.

The \ours\ evaluation covers all 9,561 scene--condition instances across the six categories. For each model, we report NAS, CPS, and NCS for route-level evaluation, LAS and MCS for lane-level evaluation, and the six category-wise contextual success rates.

\subsection{Statistical Analysis}
\label{sec:appendix_statistical_analysis}

We estimate uncertainty using scene-level bootstrap resampling. The resampling unit is the unique physical scene token instead of an individual scene--condition instance. This design preserves the dependence among multiple navigation conditions constructed from the same physical scene.

For each bootstrap replicate, 5,911 scene tokens are sampled with replacement. All scene--condition instances associated with a sampled scene are retained together. Predictions from all evaluated models associated with the same sampled scene are also retained together. This produces matched bootstrap samples across models and preserves the paired evaluation structure.

We use 10,000 scene-level bootstrap replicates to estimate uncertainty. Table~\ref{tab:ncs_confidence_intervals} reports NCS and its 95\% percentile bootstrap confidence interval for each model.

\begin{table}[t]
\centering
\caption{NCS and 95\% scene-level bootstrap confidence intervals for the 23 evaluated models.}
\label{tab:ncs_confidence_intervals}
\small
\setlength{\tabcolsep}{10pt}
\renewcommand{\arraystretch}{1.05}
\begin{tabular}{lrr}
\toprule
Model & NCS & 95\% CI \\
\midrule
OneVL & 39.30 & [34.73, 43.41] \\
ChainFlow-VLA & 35.13 & [31.19, 38.83] \\
DrivoR & 34.78 & [30.76, 38.58] \\
ReCogDrive-IL & 34.31 & [30.38, 38.01] \\
SUV & 32.03 & [27.10, 36.55] \\
ReCogDrive-RL & 31.64 & [27.82, 35.18] \\
GoalFlow & 31.54 & [27.26, 35.45] \\
NoRD & 29.51 & [25.79, 33.03] \\
WoTE & 28.52 & [24.28, 32.28] \\
AutoVLA & 27.08 & [22.93, 30.83] \\
TransFuser & 27.01 & [22.48, 31.03] \\
SparseDriveV2 & 26.60 & [22.79, 30.05] \\
DriveVLA-W0 & 16.82 & [12.11, 20.95] \\
DiffusionDrive & 14.68 & [9.92, 18.69] \\
StateMLP & 11.63 & [9.78, 13.37] \\
DP-VLA & 10.90 & [6.13, 14.88] \\
Curious-VLA & 6.02 & [0.00, 9.72] \\
CLOVER & 5.86 & [0.00, 9.45] \\
WAM-Flow & 5.86 & [0.00, 9.36] \\
SimWAM & 4.12 & [0.00, 7.41] \\
DiffusionDriveV2 & 0.00 & [0.00, 0.00] \\
Drive-JEPA & 0.00 & [0.00, 0.00] \\
DynVLA Stage 2 & 0.00 & [0.00, 0.00] \\
\bottomrule
\end{tabular}

\vspace{2pt}
\begin{minipage}{0.90\textwidth}
\scriptsize
DiffusionDriveV2, Drive-JEPA, and DynVLA Stage 2 have NAS = 0, yielding NCS = 0 in both the original evaluation and all bootstrap replicates.
\end{minipage}
\end{table}

Pairwise model comparisons are performed using the same matched bootstrap replicates. For a pair of models $a$ and $b$, let
\begin{equation}
\Delta_r^{*}
=
\mathrm{NCS}_{a}^{*(r)}
-
\mathrm{NCS}_{b}^{*(r)}
\end{equation}
denote their NCS difference in bootstrap replicate $r$.

For $B=10{,}000$ matched bootstrap replicates, the two-sided p-value is computed using an add-one correction:
\begin{equation}
p
=
\min
\left\{
1,
2
\min
\left(
\frac{
1+\sum_{r=1}^{B}
\mathbb{1}
\left[
\Delta_r^{*}\leq0
\right]
}{
B+1
},
\frac{
1+\sum_{r=1}^{B}
\mathbb{1}
\left[
\Delta_r^{*}\geq0
\right]
}{
B+1
}
\right)
\right\}.
\end{equation}

Statistical testing is performed over all
\begin{equation}
\binom{23}{2}
=
253
\end{equation}
pairwise NCS hypotheses. Benjamini--Hochberg correction is applied jointly to these 253 p-values to control the false discovery rate.

We assess uncertainty in model comparisons using matched scene-level bootstrap resampling. Of the 253 pairwise NCS comparisons, 193 are significant after Benjamini--Hochberg correction at $q<0.05$. OneVL exceeds ChainFlow-VLA by 4.17 points, with a 95\% confidence interval of [0.74, 7.43] for the paired difference and an adjusted $q$-value of 0.0252. For adjacent models ranked second through seventh, the 95\% confidence intervals for paired NCS differences all include zero.

Bootstrap confidence-interval exclusion and Benjamini--Hochberg significance are reported as two related statistical criteria. The same scene-level resampling principle is used for the confidence intervals reported in the contextual factor analysis.

\subsection{Implementation and Reproducibility Details}
\label{sec:appendix_implementation}

All reported \ours\ results use the same fixed scenario construction and evaluation rules described in Secs.~\ref{sec:appendix_scenario}, \ref{sec:appendix_evaluation}, and \ref{sec:appendix_metric}. The category definitions, geometric thresholds, path-change criteria, GT-relative suppression rules, LC continuity rules, road-compliance constraints, and metric aggregation are shared across all evaluated models.

The evaluation operates on model trajectories after conversion to a common representation of eight future ego poses over four seconds. Predicted poses are interpreted at the rear axle center before geometric evaluation. Road-relative projection, topology assignment, paired trajectory comparison, road-compliance evaluation, and NAVSIM safety evaluation are applied using the same procedure for every model.

Cases with ambiguous scene geometry are removed during benchmark construction according to Sec.~\ref{sec:appendix_scenario}. Geometry-related states produced during model evaluation are handled according to the rules in Sec.~\ref{sec:appendix_evaluation}. For category LC, predictions assigned the \textit{uncertain} continuity state are excluded from the valid denominator when computing the category success rate.

The benchmark construction and evaluation pipeline uses fixed category assignments and fixed evaluation thresholds for all reported results. Model comparisons are performed on the same scene--condition definitions under the same evaluation protocol.

\subsection{Context Factor Definitions}
\label{sec:appendix_context_factor_analysis}

The contextual factor analysis examines how in-context driving performance varies with scene structure and traffic interaction. All factor stratification is performed within each \ours\ category. This avoids directly mixing the different scene distributions of the six contextual categories.

\paragraph{Decision Urgency.}
Decision urgency characterizes how close the current scene is to requiring an immediate behavioral decision. Since the six categories represent different contextual requirements, the corresponding proxy is category specific:
\begin{equation}
u_i^{\mathrm{urg}}
=
\begin{cases}
1/t_{\mathrm{arrival}}^{\mathrm{robust}},
& k_i\in\{A,B1\}, \\[3pt]
1/d_{\mathrm{commit}},
& k_i=C, \\[3pt]
1/L_{\mathrm{parallel}}^{\mathrm{remain}},
& k_i=L, \\[3pt]
M_{\mathrm{lat}}^{\mathrm{hist}},
& k_i=LC, \\[3pt]
D_{\mathrm{GT}}^{4s}/d_{\mathrm{decision}},
& k_i=B2.
\end{cases}
\end{equation}

Here, $L_{\mathrm{parallel}}^{\mathrm{remain}}$ denotes the remaining parallel length of the target adjacent lane. $M_{\mathrm{lat}}^{\mathrm{hist}}$ denotes the magnitude of historical lateral motion used to characterize the ongoing maneuver. Larger values indicate higher decision urgency.

Within each category, urgency values are ranked and divided into Low, Medium, and High groups using category-specific rank terciles.

\paragraph{Geometric Salience.}
Geometric salience characterizes how clearly the local geometry exposes the relevant decision structure. It is constructed from the available normalized geometric components
\begin{equation}
\frac{|\theta_{\mathrm{start}}|}{15^\circ},
\qquad
\frac{|\theta_{\mathrm{branch}}^{\max}|}{45^\circ},
\qquad
\frac{d_{\mathrm{split}}}{5~\mathrm{m}},
\end{equation}
\begin{equation}
\frac{|\Delta\theta_{\mathrm{hist}}|}{30^\circ},
\qquad
\frac{|\Delta\theta_{\mathrm{lane}}|}{15^\circ},
\qquad
\frac{d_{\mathrm{lateral}}}{8~\mathrm{m}}.
\end{equation}

Each available component is clipped to the interval $[0,2]$. The geometric salience score is defined as
\begin{equation}
s_i^{\mathrm{geo}}
=
\frac{1}{|\mathcal{V}_i|}
\sum_{m\in\mathcal{V}_i}
\operatorname{clip}
\left(
g_{i,m},
0,
2
\right),
\end{equation}
where $\mathcal{V}_i$ denotes the set of available geometric components for instance $i$.

Geometric salience is therefore a composite descriptor built from the available road and motion geometry. Low, Medium, and High groups are obtained using rank terciles within each category.

\paragraph{Topological Evidence.}
Topological evidence characterizes the strength of the local structural cues associated with the contextual requirement. Its definition depends on the category.

For A and B1, the evidence is based on the relative heading of the target branch. Weak, medium, and strong evidence correspond to
\begin{equation}
|\Delta\theta_{\mathrm{target}}|
<
60^\circ,
\end{equation}
\begin{equation}
60^\circ
\leq
|\Delta\theta_{\mathrm{target}}|
<
90^\circ,
\end{equation}
and
\begin{equation}
|\Delta\theta_{\mathrm{target}}|
\geq
90^\circ.
\end{equation}

For B2, topological evidence is defined by the number of locally supported targets in the alternative direction. Weak evidence corresponds to at least two supported targets. Medium evidence corresponds to one supported target. Strong evidence corresponds to no supported target. Evidence strength therefore increases as the number of supported alternatives decreases.

For C, the evidence is based on the post-commitment progress. Weak evidence satisfies
\begin{equation}
d_{\mathrm{commit}}
<
10~\mathrm{m}.
\end{equation}
Medium evidence satisfies
\begin{equation}
10~\mathrm{m}
\leq
d_{\mathrm{commit}}
<
30~\mathrm{m}.
\end{equation}
Strong evidence satisfies
\begin{equation}
d_{\mathrm{commit}}
\geq
30~\mathrm{m}.
\end{equation}

For L, the evidence is determined by the available parallel length of the target adjacent lane. Weak, medium, and strong evidence correspond to 20--30~m, 30--60~m, and at least 60~m, respectively.

For LC, the evidence is based on historical lateral-motion magnitude. Weak evidence corresponds to less than 1.0~m. Medium evidence corresponds to 1.0--1.5~m. Strong evidence corresponds to at least 1.5~m.

The topological evidence analysis in the main paper reports Strong-minus-Weak contextual success.

\paragraph{Traffic Density.}
Traffic density is defined using a weighted count of nearby traffic participants within 40~m:
\begin{equation}
D_{\mathrm{traffic}}
=
N_{\mathrm{veh},40m}
+
0.5
\left(
N_{\mathrm{ped},40m}
+
N_{\mathrm{bike},40m}
\right)
+
1.5
N_{\mathrm{front},40m}.
\end{equation}

Here, $N_{\mathrm{veh},40m}$, $N_{\mathrm{ped},40m}$, and $N_{\mathrm{bike},40m}$ denote the numbers of vehicles, pedestrians, and bicycles within 40~m of the ego vehicle. $N_{\mathrm{front},40m}$ denotes the number of traffic participants located within the forward driving corridor.

Traffic density is divided into Low, Medium, and High groups using category-specific rank terciles.

\subsection{Traffic Interaction Comparisons}
\label{sec:appendix_traffic_interaction}

The traffic-interaction analysis compares contextual success between predefined comparison and reference conditions within the same \ours\ category.

For Lead Vehicle, the comparison condition is $\leq10$~m and the reference condition is $>25$~m. For TTC Risk, the comparison condition is $<3$~s and the reference condition is $\geq6$~s.

For Target Occupancy and Crossing Conflict, the comparison condition contains at least two actors and the reference condition contains no actors. Traffic Density compares the High group with the Low group.

All reported differences are computed as
\begin{equation}
\Delta q
=
q_{\mathrm{comparison}}
-
q_{\mathrm{reference}}.
\end{equation}

For Traffic Density, contextual success is pooled over valid model--instance predictions separately within the High and Low groups. We draw 10,000 bootstrap samples from the unique scene tokens within each category, sampling with replacement and retaining all associated scene--condition instances and model predictions together. Each replicate yields a High-minus-Low difference, from which we compute a 95\% percentile confidence interval. Asterisks in Table~\ref{tab:traffic_context_factors} indicate intervals that exclude zero.

The reported differences describe within-category associations between contextual factors and model performance. They do not assume causal effects of the corresponding traffic conditions.

\subsection{Metric Robustness Analysis}
\label{sec:appendix_metric_robustness}

We evaluate the sensitivity of model rankings to the relative weighting of NAS and CPS using
\begin{equation}
\mathrm{NCS}(w)
=
\mathrm{NAS}^{w}
\mathrm{CPS}^{1-w}.
\end{equation}

We vary $w$ from 0.30 to 0.70 in increments of 0.01 and compare each ranking with the default setting of $w=0.5$. Within [0.40, 0.60], the minimum Spearman correlation is 0.9960 and the maximum rank change is one position. Over [0.30, 0.70], these values are 0.9703 and five positions, respectively. OneVL remains the highest-ranked model at every tested weight.

We further examine sensitivity to the composition of CPS by omitting B1, B2, or C in turn and averaging the two remaining category scores. NAS remains fixed in each comparison. The resulting rankings have Spearman correlations of 0.9960, 0.9970, and 0.9970 with the default ranking, with maximum rank changes of two, one, and one positions, respectively.

\begin{table}[t]
\centering
\caption{Robustness of the \ours\ ranking under metric variations.}
\label{tab:robustness}
\small
\setlength{\tabcolsep}{5pt}
\renewcommand{\arraystretch}{1.05}
\begin{tabular}{lcc}
\toprule
Setting & \makecell{Minimum Spearman\\correlation} & \makecell{Maximum\\rank change} \\
\midrule
$w \in [0.40,0.60]$ & 0.9960 & 1 \\
$w \in [0.30,0.70]$ & 0.9703 & 5 \\
\midrule
Omit B1 from CPS & 0.9960 & 2 \\
Omit B2 from CPS & 0.9970 & 1 \\
Omit C from CPS & 0.9970 & 1 \\
\bottomrule
\end{tabular}
\end{table}

\subsection{Atomic Metric Decomposition}
\label{sec:appendix_atomic_metrics}

We further decompose the final contextual success into its underlying evaluation components. This analysis helps identify whether failures arise from alternative-condition correctness, road compliance, dynamic safety, or the paired contextual response.

Table~\ref{tab:atomic_metric_decomposition} reports the category-level decomposition pooled over all evaluable model-instance predictions. The pooled LC success rate is 85.64\%, compared with the model-averaged rate of 85.51\% in Table~\ref{tab:category_results}, because the number of evaluable LC predictions varies across models. Alt.-Condition Correctness denotes the category-specific correctness under the alternative navigation condition. It is topology-based for A and L, includes GT-relative suppression for B1, B2, and C, and corresponds to continuity correctness for LC. RoadOK denotes the full-vehicle road-compliance indicator $R_i^F$. Dynamic Safety denotes the NAVSIM safety indicator $E_i^F$.

Overall Safety requires both road compliance and dynamic safety:
\begin{equation}
\mathrm{OverallSafety}_i
=
R_i^F
\land
E_i^F.
\end{equation}

Alternative-Condition Success additionally requires the category-specific alternative-condition correctness $A_i^F$ defined in Sec.~\ref{sec:appendix_instance_success}:
\begin{equation}
q_i^{\mathrm{alt}}
=
A_i^F
\land
R_i^F
\land
E_i^F.
\end{equation}

Contextual Success further incorporates the paired contextual decision correctness $C_i$:
\begin{equation}
q_i
=
C_i
\land
R_i^F
\land
E_i^F.
\end{equation}

\begin{table*}[htbp]
\centering
\caption{Atomic metric decomposition across the six \ours\ categories.}
\label{tab:atomic_metric_decomposition}

\small
\setlength{\tabcolsep}{3.5pt}
\renewcommand{\arraystretch}{1.10}

\begin{tabular}{c|rrrrrr}
\toprule
Category
& Alt. Correct.
& RoadOK
& Dynamic Safety
& Overall Safety
& Alt. Success
& Contextual Success \\
\midrule

A
& 44.94
& 19.49
& 66.12
& 11.95
& 11.44
& 10.01 \\

B1
& 94.37
& 68.63
& 85.43
& 63.59
& 62.91
& 62.91 \\

B2
& 92.50
& 79.86
& 84.19
& 73.95
& 73.28
& 73.28 \\

C
& 88.12
& 46.87
& 67.93
& 35.85
& 35.01
& 34.98 \\

L
& 3.51
& 97.57
& 94.16
& 93.55
& 1.58
& 0.45 \\

LC
& 97.51
& 94.32
& 87.05
& 85.85
& 85.64
& 85.64 \\

\bottomrule
\end{tabular}

\vspace{2pt}
\begin{minipage}{0.98\textwidth}
\scriptsize
\textbullet\; Values are pooled over all evaluable model-instance predictions within each category.
\end{minipage}

\end{table*}

The decomposition reveals distinct failure profiles across contextual requirements. Category A is constrained by both alternative-condition correctness and road compliance. Its Alt.-Condition Correctness is 44.94\%, while RoadOK is only 19.49\%, resulting in a contextual success rate of 10.01\%. Category L shows a different pattern. RoadOK and Dynamic Safety reach 97.57\% and 94.16\%, respectively, whereas Alt.-Condition Correctness is only 3.51\%. This suggests that failures in L primarily arise from the inability to execute the feasible alternative lane intention.

Category C retains relatively high alternative-condition correctness at 88.12\%, while RoadOK and Dynamic Safety reduce the final contextual success to 34.98\%. B2 and LC achieve the highest contextual success among the six categories at 73.28\% and 85.64\%, respectively. For B1, the high Alt.-Condition Correctness does not imply that every detected behavioral response is appropriate because longitudinal adjustment and waiting can remain valid under its contextual requirement.

\subsection{Human Expert Validation}
\label{sec:appendix_human_validation}

After the scenario construction and evaluation rules were finalized, three reviewers with backgrounds in autonomous driving independently inspected the same randomly sampled subset of 500 scene--condition instances and 11,500 model-instance evaluations from all 23 evaluated models. The reviewers were blinded to the automated evaluation results during independent assessment. They subsequently discussed their judgments to reach a final consensus.

For scenario construction, the reviewers identified 3 of the 500 scene--condition instances whose assigned contextual categories were inconsistent with their consensus judgments. For model-instance evaluation, the final consensus differed from the automated evaluation outcomes for approximately 0.7\% of the 11,500 inspected model-instance evaluations.

The audit also identified ambiguity in the original NAVSIM navigation labels. Further examination of their construction provides context for this observation. NAVSIM assigns directional commands by thresholding the lateral displacement of a point 20\,m ahead along the route centerline, with left/right thresholds of $\pm2$\,m \citep{dauner2024navsim}. Its left/right commands encompass turns, lane changes, and sharp curves, leaving the maneuver type unspecified. In the inspected cases, some lane changes were associated with left/right labels, while others were not.

This ambiguity may partly contribute to the poor performance observed in category L, since the same left/right commands can correspond to different maneuver types during training.

\clearpage
\section{Qualitative Examples}
\label{sec:appendix_qualitative}

\subsection{Examples across Contextual Categories}
\label{sec:appendix_category_examples}

\begin{figure}[!ht]
    \centering
    \includegraphics[width=\linewidth]{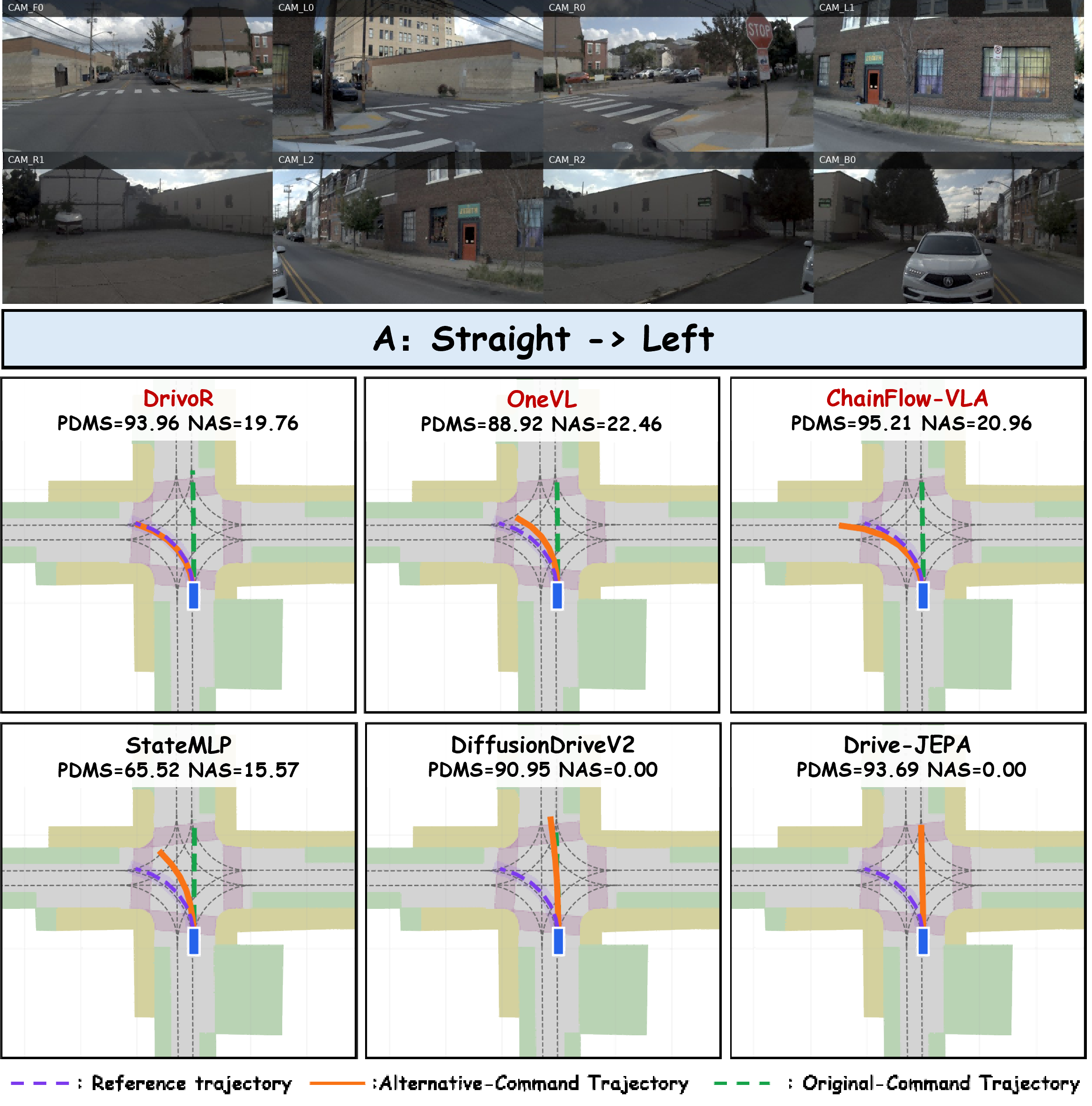}
    \caption{Category A: an immediately actionable alternative route. The navigation command changes from straight to left at an intersection where the alternative left-turn route is reachable within the evaluation horizon. The trajectories illustrate differences in how models respond to the alternative navigation intention.}
    \label{fig:qualitative_a}
\end{figure}

\begin{figure}[!ht]
    \centering
    \includegraphics[width=\linewidth]{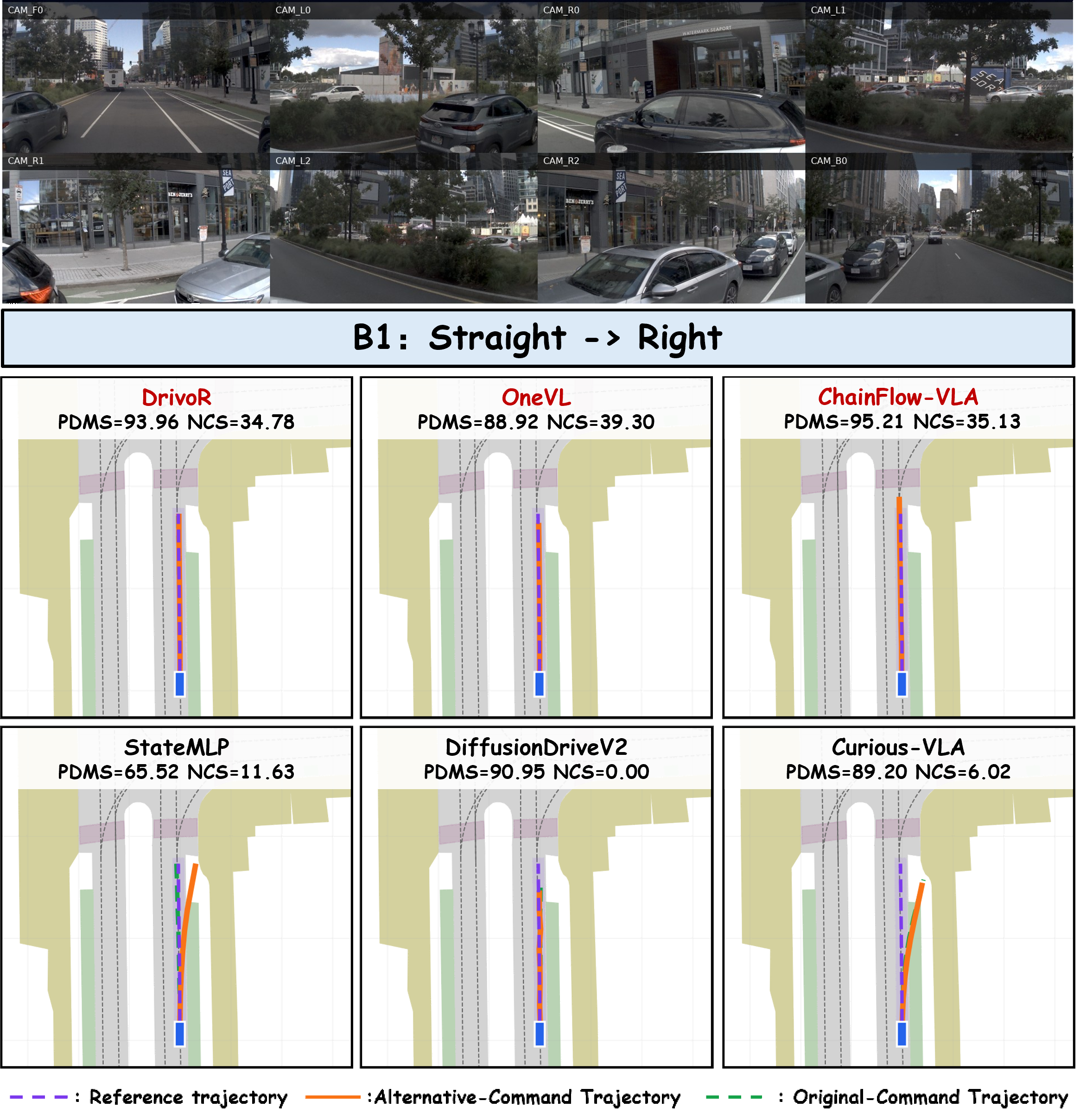}
    \caption{Category B1: an alternative route beyond the current decision window. The navigation command changes from straight to right before the right-turn target becomes actionable. Continuing along the current road remains appropriate within the evaluation horizon. The trajectories illustrate differences between maintaining the current path and deviating prematurely toward the requested direction.}
    \label{fig:qualitative_b1}
\end{figure}

\begin{figure}[!ht]
    \centering
    \includegraphics[width=\linewidth]{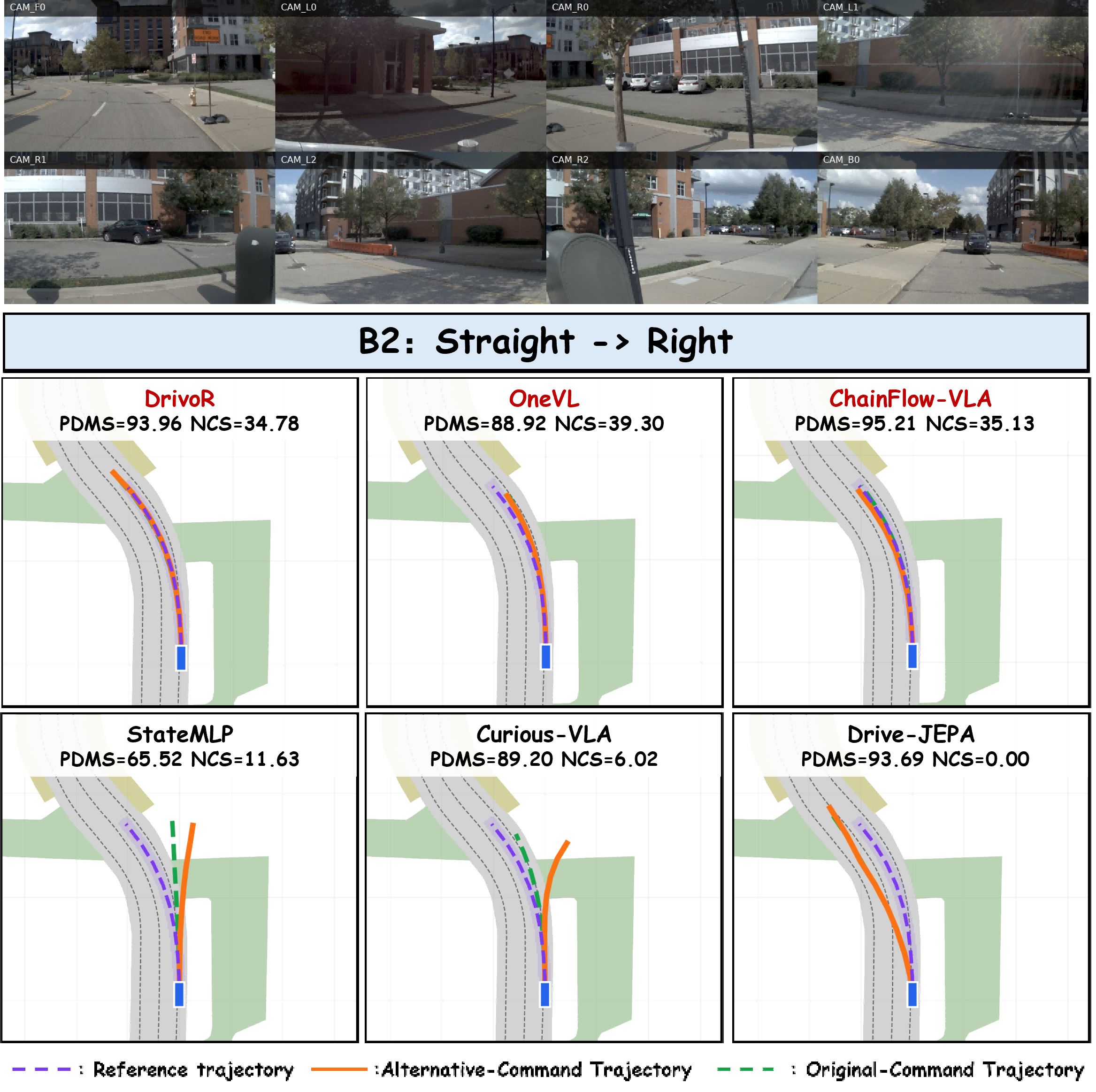}
    \caption{Category B2: a locally unsupported navigation intention. The navigation command changes from straight to right where the local road topology provides no feasible right-turn route. The trajectories illustrate differences between following the existing road geometry and deviating toward the unsupported direction.}
    \label{fig:qualitative_b2}
\end{figure}

\begin{figure}[!ht]
    \centering
    \includegraphics[width=\linewidth]{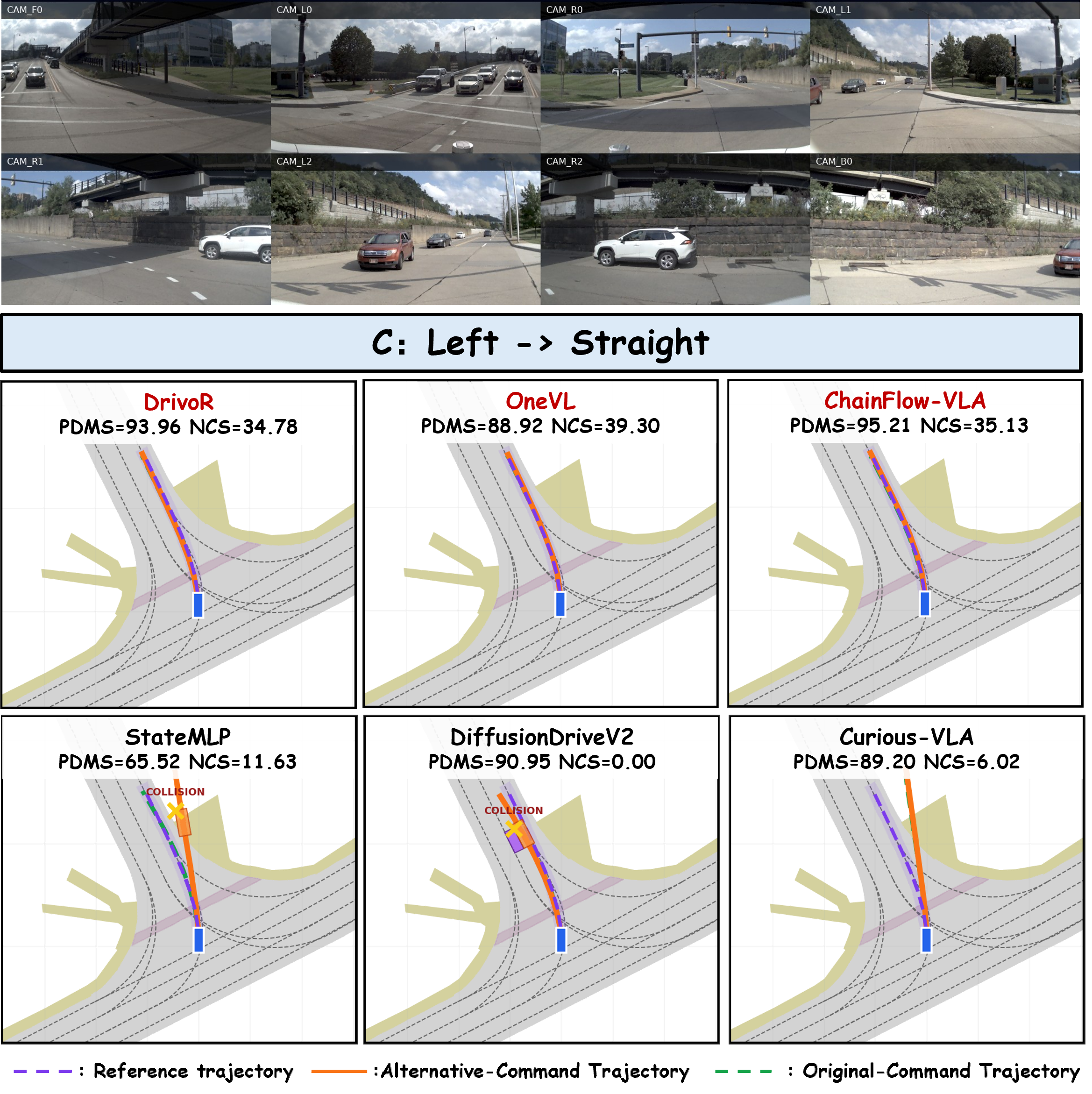}
    \caption{Category C: a navigation intervention after route commitment. The navigation command changes from left to straight after the ego vehicle has passed the route commitment point. The trajectories illustrate differences between continuing along the committed route and deviating in response to the changed command. Collision annotations identify the marked failures in the displayed predictions.}
    \label{fig:qualitative_c}
\end{figure}

Figures~\ref{fig:qualitative_a}--\ref{fig:qualitative_c} present qualitative examples from categories A, B1, B2, and C to illustrate model responses to navigation interventions under different driving contexts. Each figure shows the shared camera observations and trajectories predicted under the original and alternative navigation commands. The displayed scores summarize model performance over the corresponding evaluation sets.

\clearpage

\clearpage

\clearpage

\subsection{Examples across Models}
\label{sec:appendix_model_examples}

We present additional qualitative examples for ten evaluated models in Figures~\ref{fig:model_onevl}--\ref{fig:model_dynvla}. Each figure contains three selected scene--condition instances, with camera views and corresponding bird's-eye-view visualizations of the predicted trajectories. Green denotes predictions under the original navigation command, while yellow in the camera views and orange in the bird's-eye views denote predictions under the alternative command. The examples illustrate route choices, responses to navigation interventions, and safety violations across different driving contexts. The displayed mPDMS and NCS values summarize model performance over the corresponding evaluation sets.

\clearpage
\begingroup
\setlength{\intextsep}{6pt}
\setlength{\floatsep}{10pt}
\captionsetup{skip=5pt}

\begin{figure}[!ht]
    \centering
    \includegraphics[width=\linewidth]{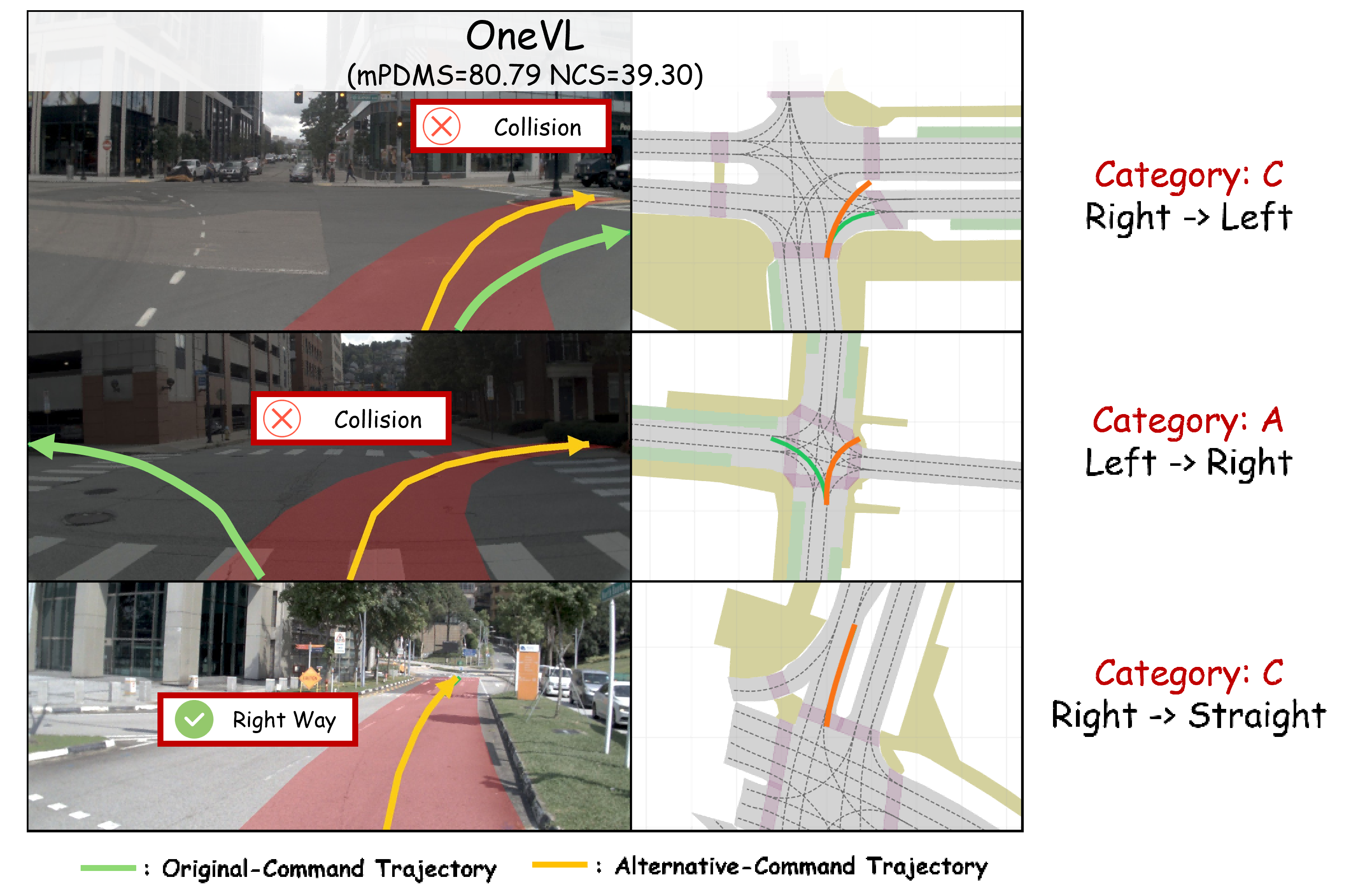}
    \caption{Qualitative examples for OneVL. The first two cases show collisions following navigation interventions in categories C and A. The third case shows a directionally correct response in category C.}
    \label{fig:model_onevl}
\end{figure}

\begin{figure}[!ht]
    \centering
    \includegraphics[width=\linewidth]{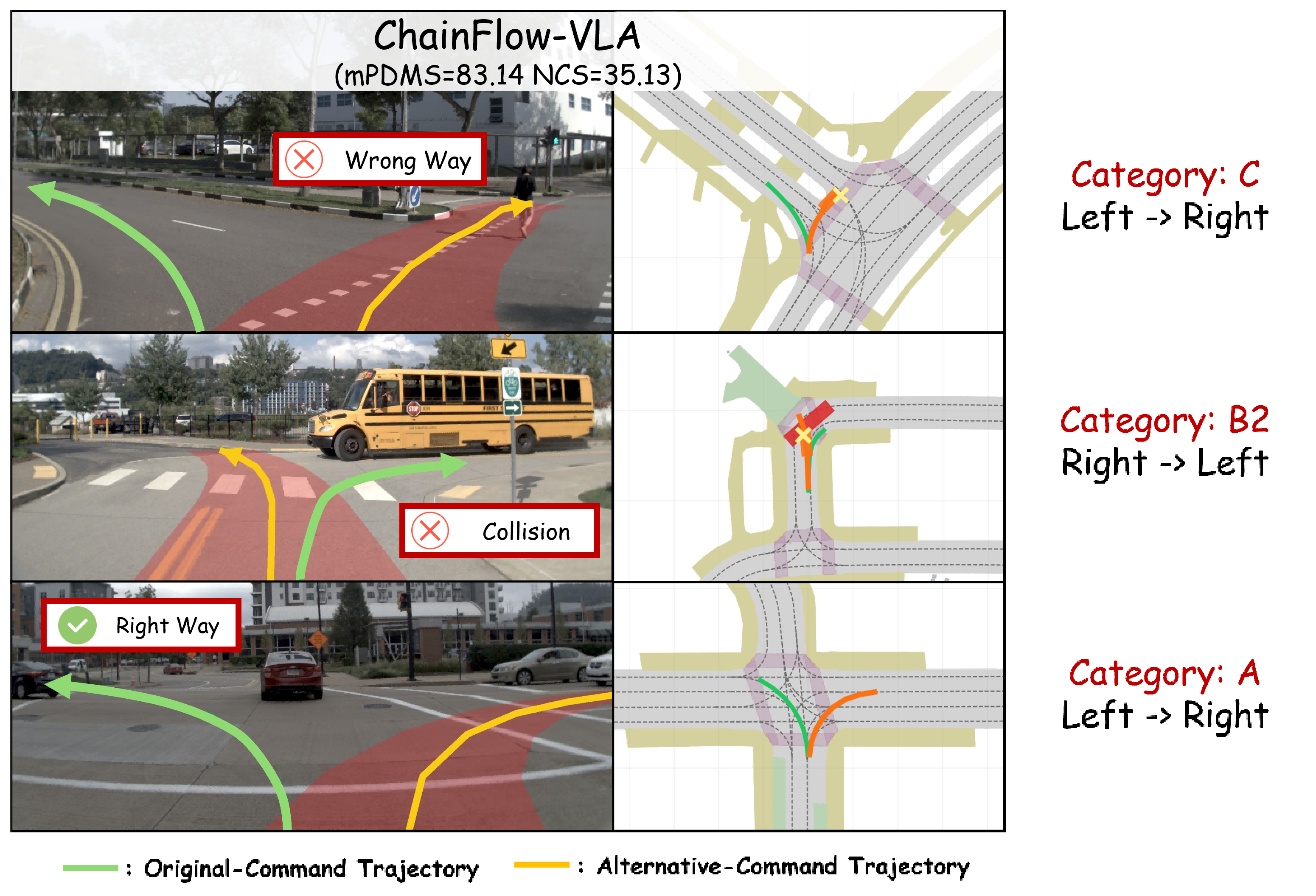}
    \caption{Qualitative examples for ChainFlow-VLA. The cases show an incorrect route choice after route commitment in category C, a collision in category B2, and a correct alternative route choice in category A.}
    \label{fig:model_chainflow}
\end{figure}

\clearpage
\begin{figure}[!ht]
    \centering
    \includegraphics[width=\linewidth]{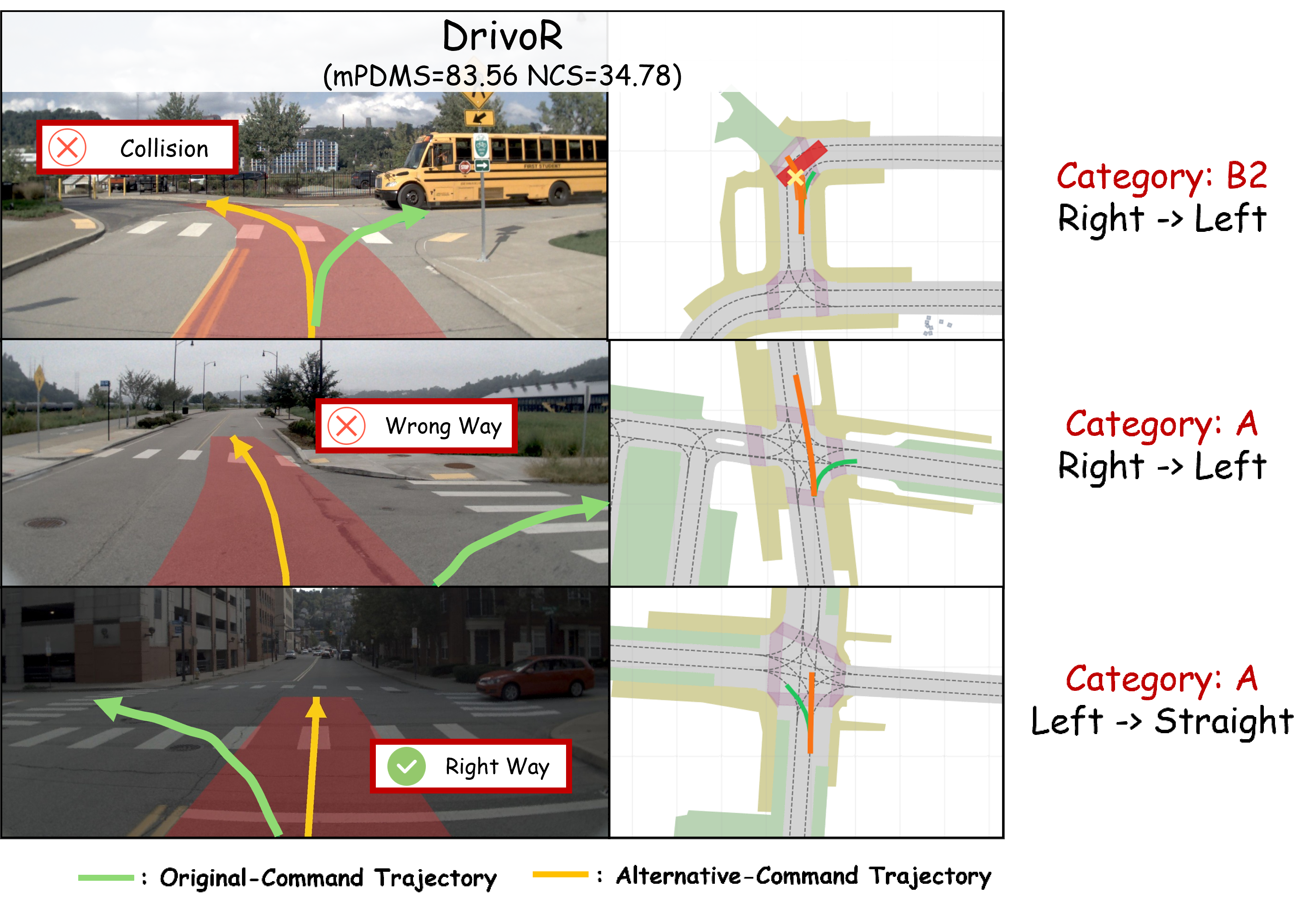}
    \caption{Qualitative examples for DrivoR. The first case shows a collision in category B2. The two category A cases illustrate different outcomes when responding to an immediately actionable alternative navigation intention, including an incorrect route choice and a correct route choice.}
    \label{fig:model_drivor}
\end{figure}

\begin{figure}[!ht]
    \centering
    \includegraphics[width=\linewidth]{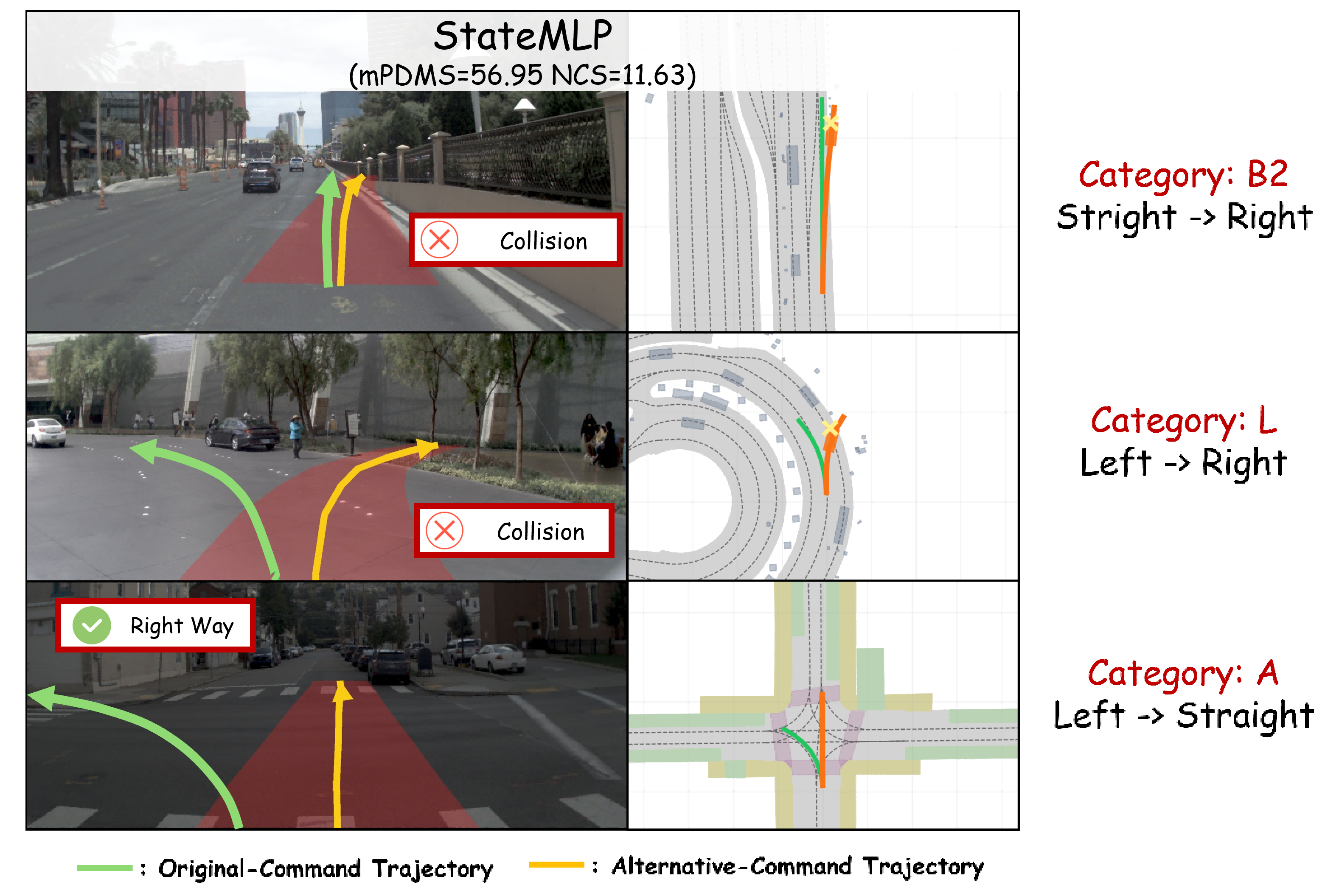}
    \caption{Qualitative examples for StateMLP. The first two cases show collisions in categories B2 and L. The third case shows a correct alternative route choice in category A.}
    \label{fig:model_statemlp}
\end{figure}

\clearpage
\begin{figure}[!ht]
    \centering
    \includegraphics[width=\linewidth]{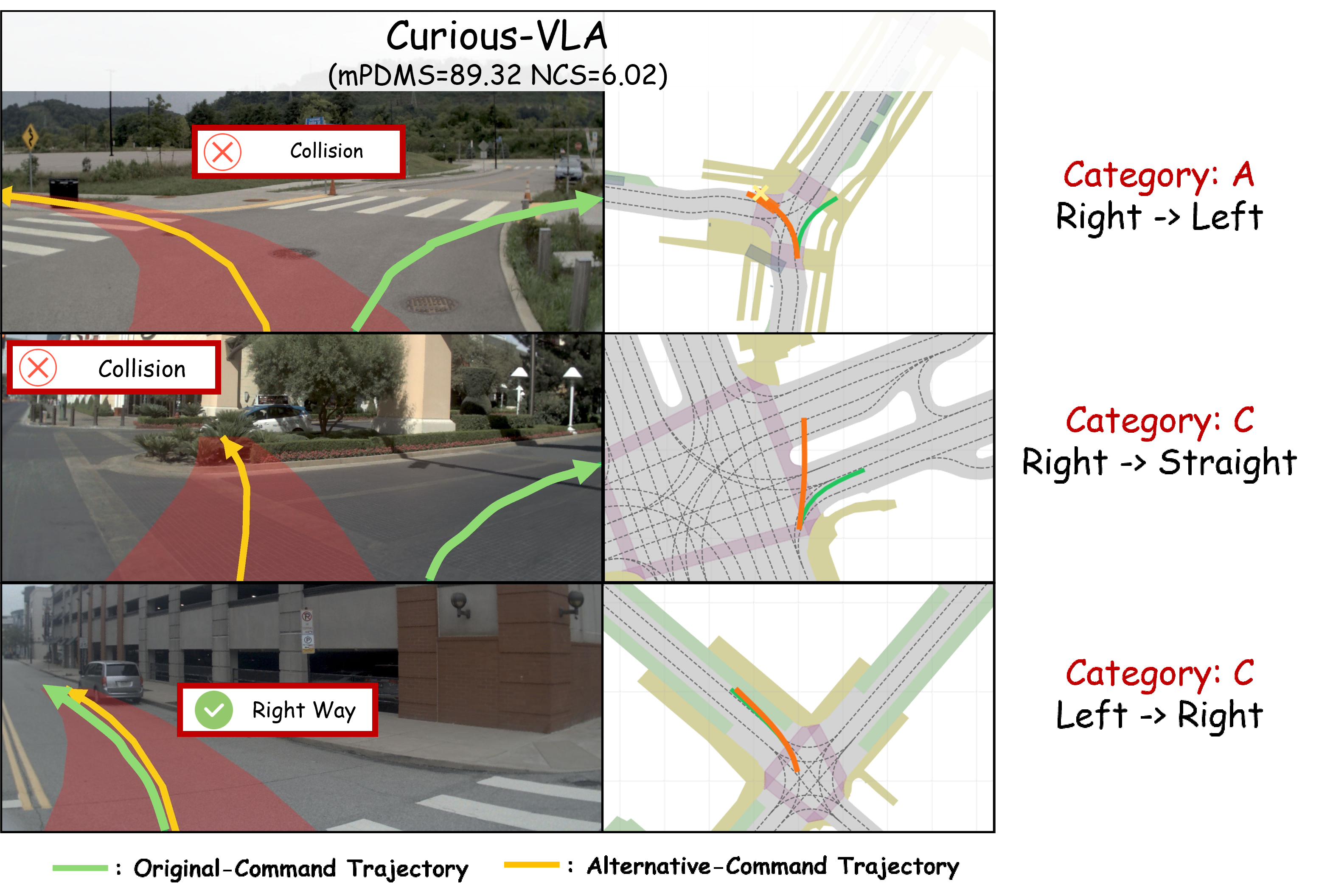}
    \caption{Qualitative examples for Curious-VLA. The first two cases show collisions in categories A and C. The third case shows continued travel along the committed route after the navigation command changes in category C.}
    \label{fig:model_curious_vla}
\end{figure}

\begin{figure}[!ht]
    \centering
    \includegraphics[width=\linewidth]{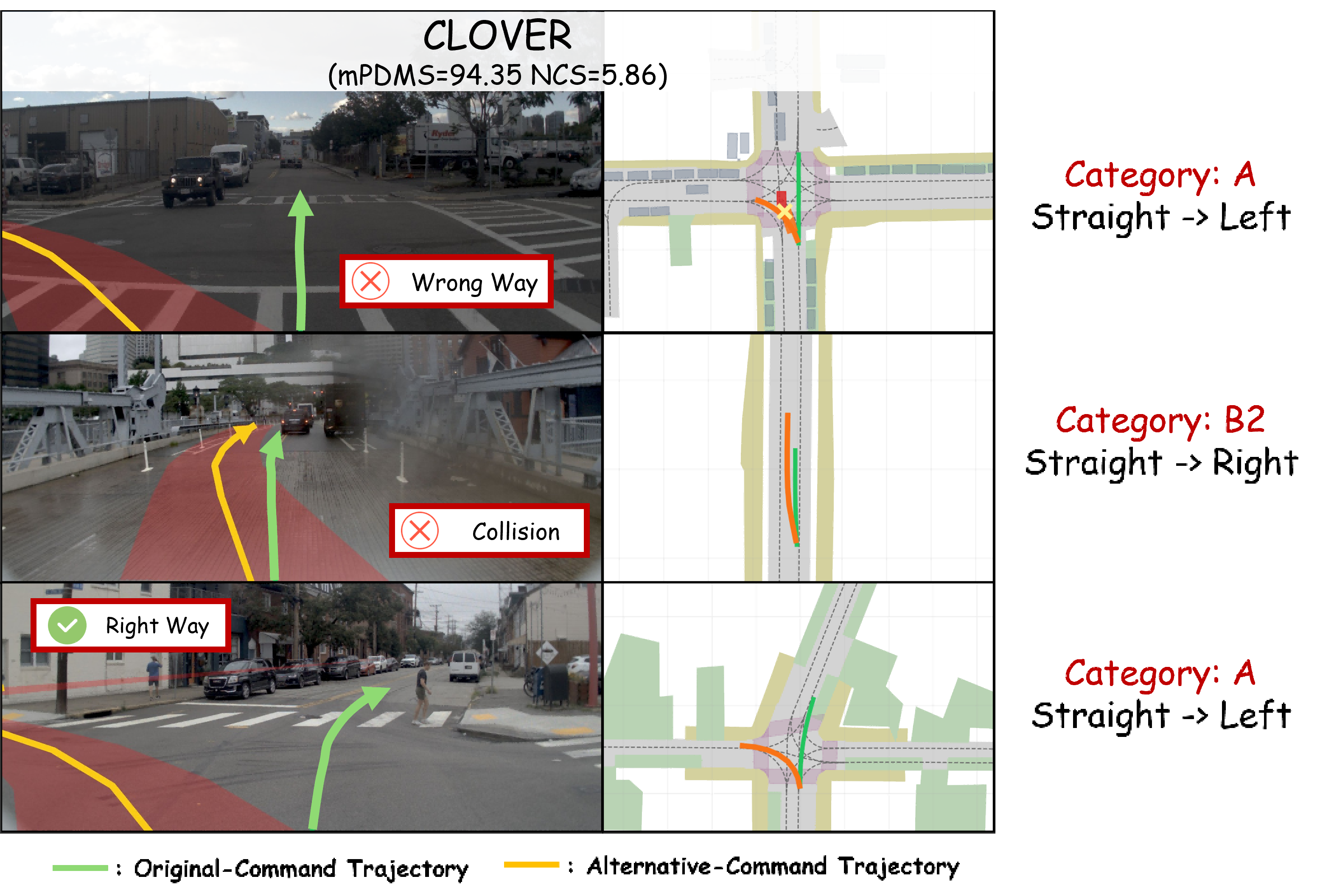}
    \caption{Qualitative examples for CLOVER. The cases show an incorrect alternative route choice in category A, a collision in category B2, and a correct alternative route choice in another category A scene.}
    \label{fig:model_clover}
\end{figure}

\clearpage
\begin{figure}[!ht]
    \centering
    \includegraphics[width=\linewidth]{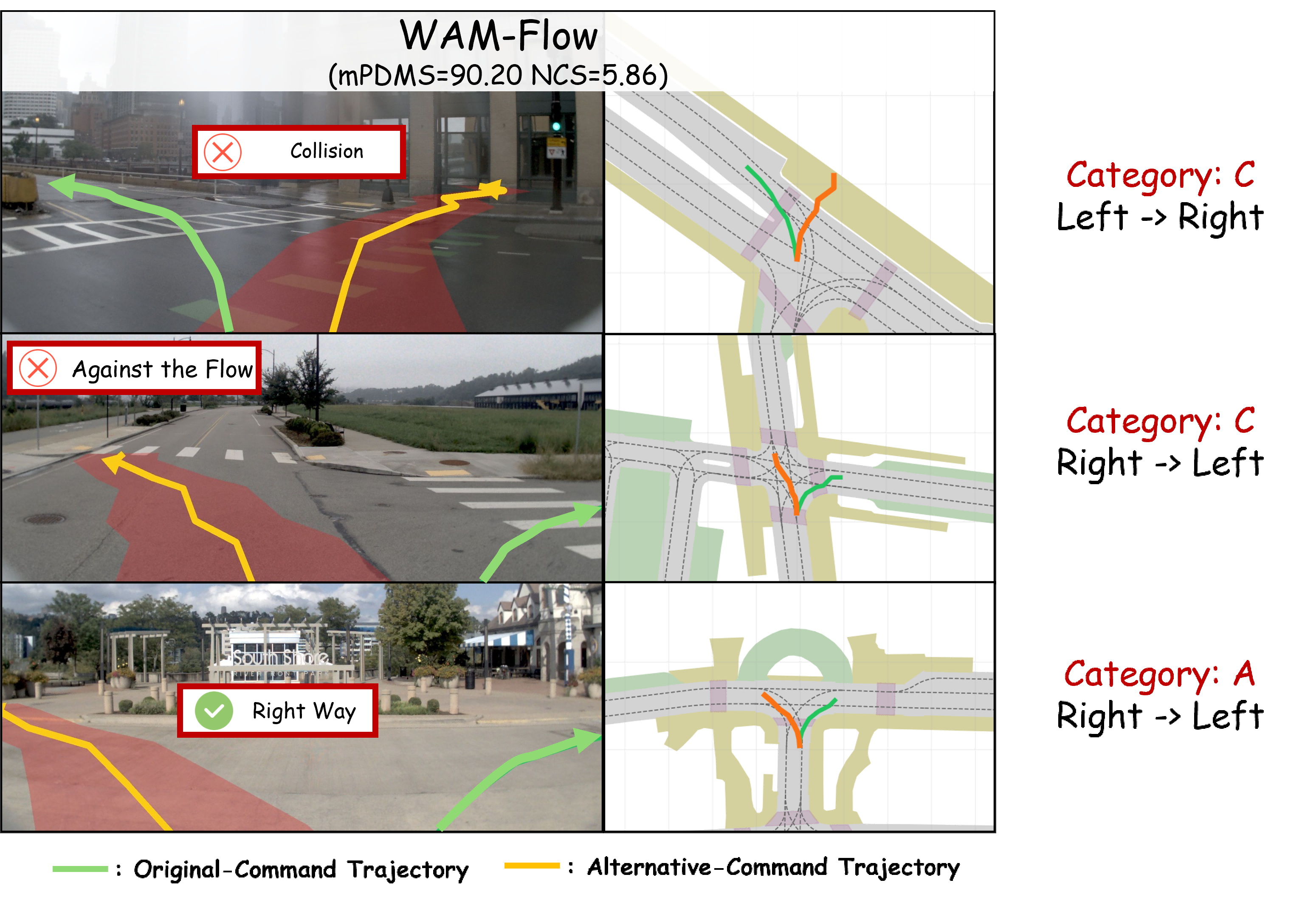}
    \caption{Qualitative examples for WAM-Flow. The two category C cases show a collision and travel against the traffic direction following navigation interventions after route commitment. The category A case shows a correct alternative route choice.}
    \label{fig:model_wam_flow}
\end{figure}

\begin{figure}[!ht]
    \centering
    \includegraphics[width=\linewidth]{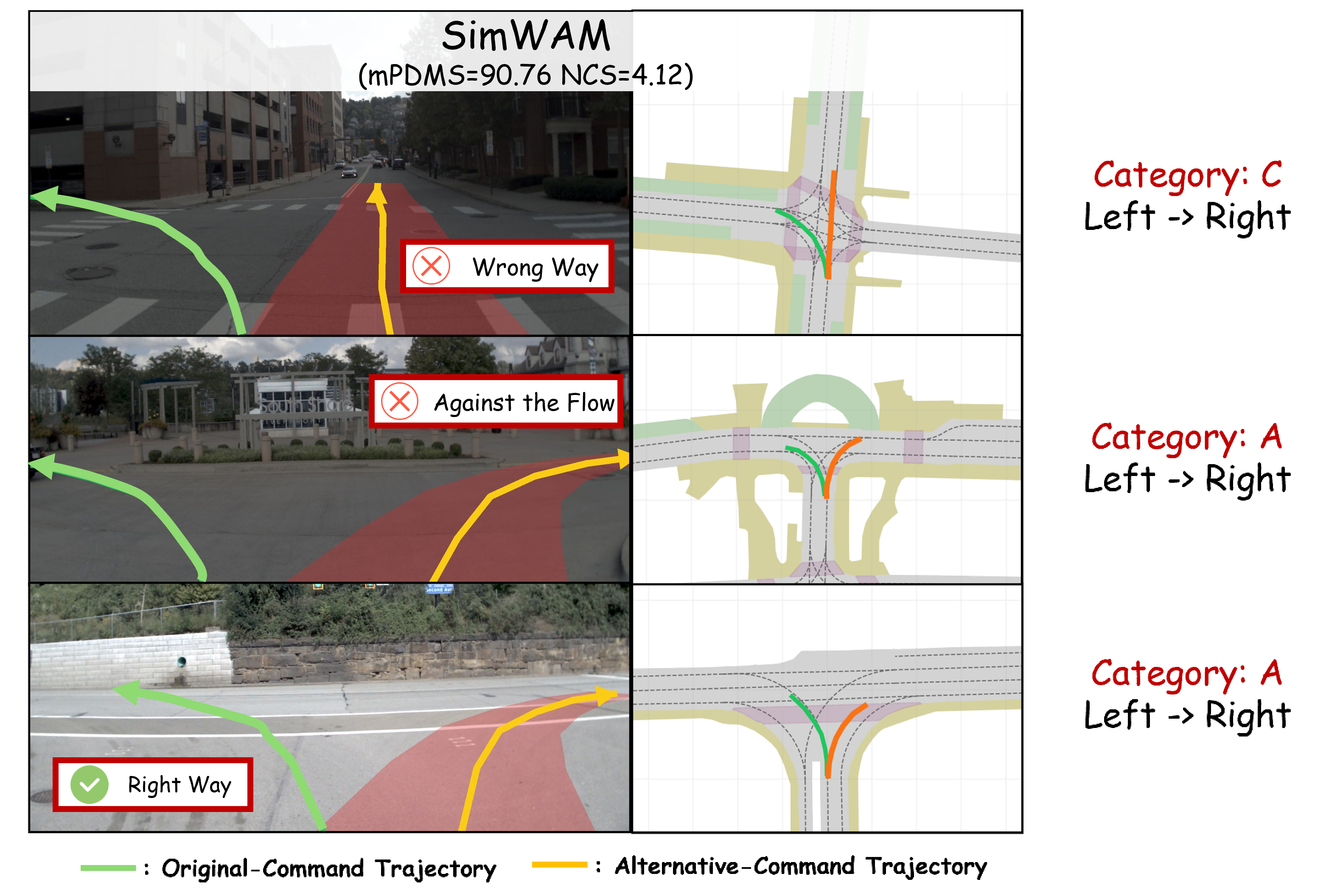}
    \caption{Qualitative examples for SimWAM. The first case shows an incorrect route choice in category C. The two category A cases show travel against the traffic direction and a correct alternative route choice, respectively.}
    \label{fig:model_simwam}
\end{figure}

\clearpage
\begin{figure}[!ht]
    \centering
    \includegraphics[width=\linewidth]{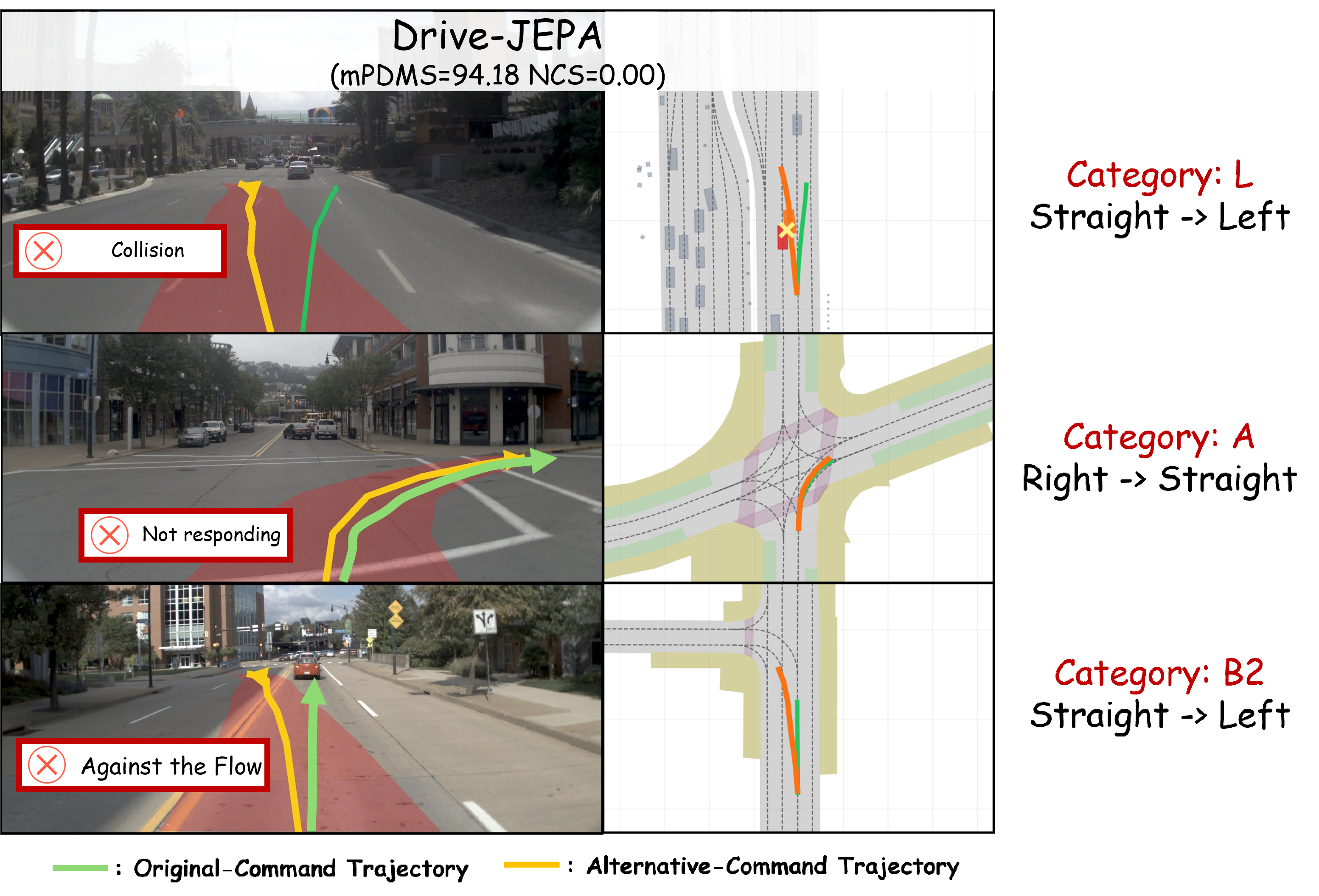}
    \caption{Qualitative examples for Drive-JEPA. The cases show a collision in category L, little visible change between the paired predictions in category A, and travel against the traffic direction in category B2.}
    \label{fig:model_drive_jepa}
\end{figure}

\begin{figure}[!ht]
    \centering
    \includegraphics[width=\linewidth]{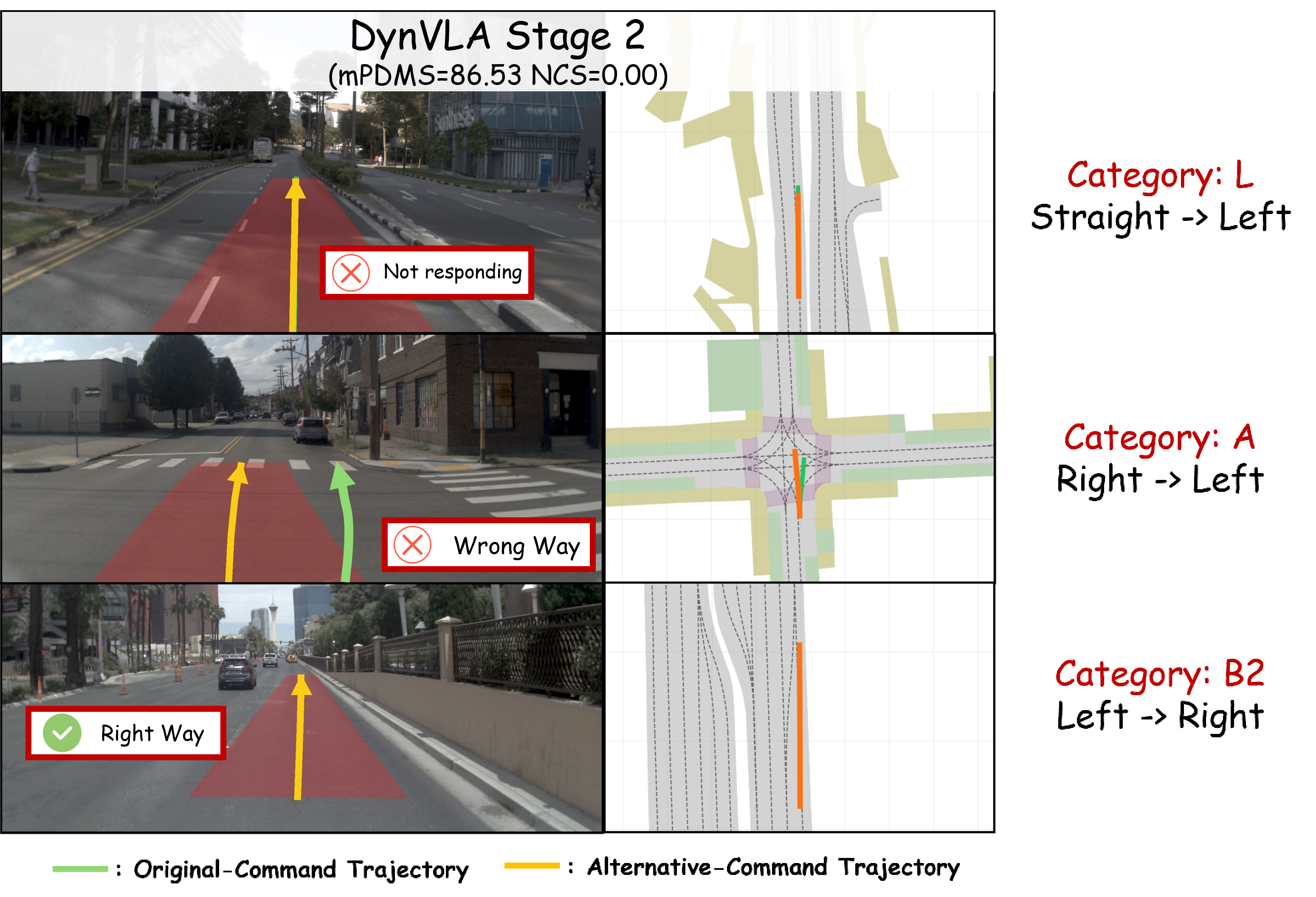}
    \caption{Qualitative examples for DynVLA Stage 2. The cases show little visible response to a lane-level navigation intervention in category L, an incorrect route choice in category A, and continued travel along the current road in category B2.}
    \label{fig:model_dynvla}
\end{figure}

\clearpage
\endgroup

\end{document}